\documentclass{ps}

\usepackage{amscd,amssymb,graphics}

\newcommand{\R}{{\mathbb R}}
\newcommand{\I}{{\mathbb I}}
\newcommand{\C}{{\mathbb C}}
\newcommand{\supp}{{\mathrm{supp}\,}}
\newcommand{\ve}{{\varepsilon}}

\newcommand{\abs}[1]{\lvert#1\rvert}

\newcommand{\e}{{\epsilon}}
\newcommand{\N}{{\mathbb N}}
\newcommand{\Q}{{\mathbb Q}}

\newcommand{\E}{{\mathbb E}}
\newcommand{\Hg}{{\mathbb H}}

\begin{document}

\title{Universal consistency of the $k$-NN rule in metric spaces and Nagata dimension. II}
\thanks{S.K. was supported by Japan Society for the Promotion of Science (JSPS) KAKENHI Grant Number 19K20347 and KIXOXIA encouragement research grant at the earlier stages of this work.}
\thanks{V.G.P. was supported by the DCR-A fellowship 300050/2022-4 of the Program of Scientific and Technological Development of the State of Para\'\i ba, Brazil by CNPq and FAPESQ.}
\runningtitle{$k$-NN rule in metric spaces. II}

\author{Sushma Kumari}\address{Defense Institute of Advanced Technology (DIAT), Pune, Maharashtra, 411025 India  {\tt   sushmak@diat.ac.in}}
\author{Vladimir G. Pestov}\address{Departamento de Matem\'atica, Universidade Federal da Para\'\i ba, Jo\~ao Pessoa, PB, Brasil}
\secondaddress{Department of Mathematics and Statistics,
       University of Ottawa, 
       Ottawa, ON K1N 6N5, Canada {\tt vpest283@uOttawa.ca}}
\date{Version as of March 20, 2024}
\begin{abstract} 
We continue to investigate the $k$ nearest neighbour ($k$-NN) learning rule in complete separable metric spaces. Thanks to the results of C\'erou and Guyader (2006) and Preiss (1983), this rule is known to be universally consistent in every such metric space that is sigma-finite dimensional in the sense of Nagata. Here we show that the rule is strongly universally consistent in such spaces in the absence of ties. Under the tie-breaking strategy applied by Devroye, Gy\"{o}rfi, Krzy\.{z}ak, and Lugosi (1994) in the Euclidean setting, we manage to show the strong universal consistency in non-Archimedian metric spaces (that is, those of Nagata dimension zero). Combining the theorem of C\'erou and Guyader with results of Assouad and Quentin de Gromard (2006), one deduces that the $k$-NN rule is universally consistent in metric spaces having finite dimension in the sense of de Groot. In particular, the $k$-NN rule is universally consistent in the Heisenberg group which is not sigma-finite dimensional in the sense of Nagata as follows from an example independently constructed by Kor\'anyi and Reimann (1995) and Sawyer and Wheeden (1992).
 \end{abstract}
\maketitle

\section*{Introduction}

The problem of describing those (separable, complete) metric spaces in which the $k$ nearest neighbour classifier is universally (weakly) consistent still remains open. The same applies to the strong universal consistency under some reasonable tie-breaking strategy. In this paper, we are motivated by those two problems and closely related questions.

The main tool in this direction is the theorem by C\'erou and Guyader \cite{CG}, who have shown that the $k$-NN classifier is (weakly) consistent under the assumption that the regression function $\eta(x)$ satisfies the weak Lebesgue--Besicovitch differentiation property. While it is unknown if this property actually follows from the consistency of the $k$-NN classifier, it is now possible to deduce the  universal consistency for every metric space having the weak Lebesgue--Besicovitch property for every probability measure. A large class of such metric spaces was previously isolated by Preiss \cite{preiss83}: the so-called sigma-finite dimensional metric spaces in the sense of Nagata \cite{nagata,ostrand}. Thus, it follows that in every separable metric space that is sigma-finite dimensional in the sense of Nagata the $k$-NN classifier is universally consistent. In the part I of this work \cite{CKP}, we have given a direct proof of the result in the spirit of the original argument of Stone for Euclidean spaces \cite{stone:77}, illustrating the similarities and differences of the argument in this more general setting.

One observation of the present paper is that the conclusion of the result holds for a strictly more general class of metric spaces. Assouad and Quentin de Gromard have shown \cite{AQ} that the Lebesgue--Besicovitch differentiation property is true for metric spaces that are finite dimensional in the sense of de Groot. In particular, modulo the results of \cite{CG}, the $k$-NN classification rule is universally consistent in such spaces. Among the most studied examples of such metric spaces is the Heisenberg group $\Hg$. It is known that the Heisenberg group has infinite Nagata dimension (this was shown independently by Kor\'anyi and Reimann \cite{KR} and Sawyer and Wheeden \cite{SW}). In fact, their argument also implies that $\Hg$ is not sigma-finite dimensional in the sense of Nagata. Thus, the $k$-NN classifier is universally consistent in the Heisenberg group, and the property of being sigma-finite dimensional in the sense of Nagata is not a necessary condition. This observation, the subject of Section \ref{s:degroot}, refutes the conjecture made by us in part I \cite{CKP}. 

It is also noteworthy that the example of the Heisenberg group answers in the negative a question asked by Preiss in 1983 \cite{preiss83}: suppose a metric space $\Omega$ satisfies the Lebesgue--Besicovitch differentiation property for every sigma-finite locally finite measure, will it satisfy the strong Lebesgue--Besicovitch differentiation property for every such measure too? While this must be well-known to the experts, we are unaware of this being mentioned explicitly anywhere. 

In the remaining part of the article we proceed to the strong universal consistency of the $k$-NN classifier in metric spaces. In Section \ref{s:noties} we show that in the absence of distance ties, the $k$-NN rule is strongly universally consistent in every separable sigma-finite dimensional space in the sense of Nagata. The argument follows closely the proof in the Euclidean case belonging originally to Devroye and Gy\"orfi \cite{DG} and Zhao \cite{Z} as presented in the book \cite{DGL} (Theorem 11.1). Clearly, the key geometric lemma using Nagata dimension is a bit different. 
Section \ref{s:noties} is a revised version of a part of the PhD thesis of the first-named author \cite{kumari}.

Adopting a specific paradigm of uniform tie-breaking belonging to Devroye, Gy\"{o}rfi, Krzy\.{z}ak, and Lugosi \cite{DGKL} who applied it in the Euclidean case, we show that the $k$-NN classifier is strongly universally consistent in the non-Archimedian metric spaces, that is, those satisfying the strong triangle inequality: $d(x,z)\leq\max\{d(x,y),d(y,z)\}$. The same holds in a slightly more general class of metric spaces of Nagata dimension zero. We were unable to extend the result to all (sigma) finite dimensional metric spaces in the sense of Nagata, but already the non-Archimedian case is, we believe, important, as it is, intuitively, where the distance ties occur most often. It is worth noting that a direct analogue of a crucial technical geometric lemma proved in \cite{DGKL} in the Euclidean case fails in non-Archimedian metric spaces with measure, revealing a rather interesting difference in their underlying geometries. This is the subject of our Section \ref{zero}.

In the concluding short section \ref{s:conjecture} we propose a new version of the conjecture aimed to describe those complete separable metric spaces in which the $k$-NN classifier is universally consistent.

\section{Preliminaries: learning rules\label{s:rules}}

\subsection{Learning in a measurable space\label{subs:learning}}
Let $\Omega=(\Omega,{\mathcal A})$ be a measurable space, that is, a non-empty set $\Omega$ equipped with a sigma-algebra of subsets $\mathcal A$. The product $\Omega\times \{0,1\}$ becomes a measurable space in a natural way. The elements $x\in \Omega$ are known as {\em unlabelled points}, and elements $(x,y)\in \Omega\times \{0,1\}$ are {\em labelled points.}
A finite sequence of labelled points, $\sigma=(x_1,x_2,\ldots,x_n,y_1,y_2,\ldots,y_n)\in \Omega^n\times \{0,1\}^n$, is a {\em labelled sample.} Here it is probably important to stress that a sample is a sequence and not a subset, as it may have repetitions. 

A {\em classifier} in $\Omega$ is a mapping
\[T\colon \Omega\to \{0,1\},\]
assigning a label to every point. The mapping is usually assumed to be measurable (or, more generally, universally measurable, that is, measurable with regard to the intersection of all possible completions of the sigma-algebra). 
This assumption is necessary in order for things like the misclassification error to be well defined, although some authors are allowing for non-measurable maps, working with the outer measure instead.

Let $\tilde\mu$ be a probability measure defined on the measurable space $\Omega\times \{0,1\}$. Denote $(X,Y)$ a random element of $\Omega\times \{0,1\}$ following the law $\tilde\mu$. The misclassification error of a classifier $T$ is the quantity
\begin{align*}
\mbox{err}_{\tilde\mu}(T) &= \tilde\mu \{(x,y)\in \Omega\times \{0,1\}\colon T(x)\neq y\} \\
&= P[T(X)\neq Y].
\end{align*}
The misclassification error cannot be smaller than the {\em Bayes error,} which is the infimum (in fact, the minimum) of the errors of all the classifiers $T$ defined on $\Omega$:
\[\ell^{\ast}=\ell^{\ast}_{\tilde\mu}=\inf_{T}{\mathrm{err}}_{\tilde\mu}(T).\]

A ({\em supervised binary classification}) {\em learning rule} in $(\Omega,{\mathcal A})$ is a mapping, $g$, that, when shown a labelled sample, $\sigma$, produces a classifier, $g(\sigma)$. In other words, a learning rule determines a label of each point $x$ on the basis of a labelled learning sample $\sigma$:
\[{g}\colon\bigcup_{n=1}^{\infty}\Omega^n\times \{0,1\}^n\times \Omega \ni (\sigma,x)\mapsto {g}(\sigma)(x) \in \{0,1\}.\]
Again, the map above is usually assumed to be (universally) measurable with regard to the natural sigma-algebra generated by $\mathcal A$ through the finite products and then countable unions. 

We denote the restriction of $g$ to $\Omega^n\times \{0,1\}^n$ by $g_n$. This way, one can think of a learning rule $g$ as a sequence of maps and write $g=(g_n)$.

The labelled datapoints are modelled by a sequence of independent, identically distributed random elements $(X_n,Y_n)\in \Omega\times \{0,1\}$ following the law $\tilde\mu$. For each $n$, the {\em misclassification error} of the rule $g$ restricted to $\Omega^n\times \{0,1\}^n$, that is, ${g}_n$, is the random variable 
\[\mbox{err}_{\tilde\mu}{g}_n=\mbox{err}_{\tilde\mu}{g}_n(D_n), \]
where $D_n$ is a random labelled $n$-sample, $D_n=(X_1,Y_1,X_2,Y_2,\ldots,X_n,Y_n)$. 

Define the measure $\mu = \tilde\mu\circ\pi^{-1}$, where $\pi$ is the first coordinate projection of $\Omega\times \{0,1\}$. This is a probability measure on $(\Omega,{\mathcal A})$. Now define a finite measure $\mu_1$ on $\Omega$ by $\mu_1(A)=\tilde\mu(A\times\{1\})$. Clearly, $\mu_1$ is absolutely continuous with regard to $\mu$. Define the {\em regression function}, $\eta\colon \Omega\to [0,1]$, as the corresponding Radon--Nikod\'ym derivative
\begin{align*}
\eta(x) & =\frac{d\mu_1}{d\mu} \\
&= P[Y=1\mid X=x],
\end{align*}
that is, the conditional probability for $x$ to be labelled $1$. (For the Radon--Nikod\'ym theorem in our abstract setting, see \cite{fremlin2}, 232E and 232B.) 

Notice that since the regression function $\eta$, together with the measure $\mu$, allows to fully reconstruct the measure $\tilde\mu$, a learning problem in a measurable space $(\Omega,{\mathcal A})$ can be alternatively given either by the measure $\tilde\mu$ or by the pair $(\mu,\eta)$. We will sometimes denote the corresponding Bayes error by $\ell^{\ast}_{\mu,\eta}$.

Given a classifier $T=\chi_C$, the misclassification error can be written as
\begin{equation}
\mbox{err}_{\tilde\mu}(T) = \int_C(1-\eta)\,d\mu + \int_{\Omega\setminus C}\eta\,d\mu.
\label{eq:expressionforBayes}
\end{equation}

Now it is easy to see that the Bayes error $\ell^\ast=\ell^{\ast}_{\mu,\eta}$ is achieved at exactly those classifiers $T$ satisfying
\[T(x) = \begin{cases} 1,&\mbox{ for $\mu$-almost all }x\mbox{ such that }\eta(x)>\frac 12,\\
0,&\mbox{ for $\mu$-almost all }x\mbox{ such that }\eta(x)<\frac 12.
\end{cases}
\]
(At the points where $\eta$ equals $1/2$, the value of a Bayes classifier -- or any classifier -- does not affect the error.)
Such classifiers are known as {\em Bayes classifiers.}

A rule $g$ is {\em consistent} (or {\em weakly consistent}) under $\tilde\mu$ if
\[\mbox{err}_{\tilde \mu}{g}_n \overset{p}{\to} \ell^{\ast}_{\tilde\mu},\]
where the convergence is in probability, 
and {\em universally consistent} if $g$ is consistent under every probability measure $\tilde\mu$ on $\Omega\times \{0,1\}$. In this paper, {\em consistency} will be synonymous with {\em weak consistency}.

In a similar way, one defines the strong consistency. A {\em labelled sample path} is an infinite sequence of i.i.d. elements of $\Omega\times\{0,1\}$ each one following the law $\tilde\mu$.
A rule $g$ is {\em strongly consistent} under $\tilde\mu$ if 
\[\mbox{err}_{\tilde \mu}{g}_n(D_n) \to \ell^{\ast}_{\tilde\mu},\]
where the convergence is along almost every infinite labelled sample path $D_{\infty}=(X_1,Y_1),(X_2,Y_2),\ldots$, and $D_n$ denotes the initial segment of the path $D_{\infty}$. A rule is {\em strongly universally consistent} if it is strongly consistent under every probability measure on the space of labelled points. Clearly, strong consistency implies consistency. 

Recall that the {\em Borel sigma-algebra} (or {\em Borel structure}) of a topological space $\Omega$ is the smallest sigma-algebra containing all open sets. In particular, every metric on a set generates a Borel sigma-algebra.
A {\em standard Borel space} is a set equipped with a sigma-algebra that is the Borel sigma-algebra generated by some complete separable metric.
The usual setting for statistical learning is a standard Borel space as $\Omega$. This will be the setting for our paper as well. However, apriori there are no restrictions for studying learning problems in more general measurable spaces.

\subsection{The $k$ nearest neighbour classification rule}

Let now $\Omega$ be a metric space.
The $k$-NN classifier in $\Omega$ is a learning rule, defined by selecting the label ${g}_n(\sigma)(x)\in\{0,1\}$ for a point $x$ on the basis of a labelled $n$-sample $\sigma=\sigma_n=(x_1,x_2,\ldots,x_n,y_1,y_2,\ldots,y_n)$, $x_i\in\Omega$, $y_i\in\{0,1\}$, by the majority vote among the values of $y_i$ corresponding to the $k=k_n$ nearest neighbours of $x$ in the learning sample $\sigma$. 

There is an issue of possibly occurring ties, which come in two types. One is the voting tie, when $k$ is even and we may have a split vote. This can be broken, in fact, in any way, without affecting the consistency of the classifier. For instance, in such cases one can always choose the label $1$ (as we do below), or just assign the label in a random way. Or else one can only work with odd values of $k_n$.

It may also be that there are more than $k$ nearest neighbours of $x$ within $\sigma$ that are at the same distance. This requires a tie-breaking rule. 
Given $k$ and $n\geq k$, define 
\begin{equation}
r^{\sigma_n}_{k\mbox{\tiny -NN}}(x)=\min\{r\geq 0\colon \sharp\{i=1,2,\ldots,n\colon x_i\in \bar B(x,r)\}\geq k\}.
\label{eq:rknn}
\end{equation}
In other words, this is the smallest radius of a closed ball around $x$ containing at least $k$ nearest neighbours of $x$ in the sample $\sigma_n$.

A {\em $k$ nearest neighbour mapping} is a function 
\[N_k\colon\Omega^n\times\Omega\to\Omega^k\]
which, given an unlabelled $n$-sample $\sigma$ and a point $x$, selects a $k$-subsample $N^{\sigma}_k(x) \sqsubset \sigma$ so that 
\begin{enumerate}
\item
all elements of $N_k^{\sigma}(x)$ are at a distance $\leq r^{\sigma_n}_{k\mbox{\tiny -NN}}(x)$ from $x$, and
\item
all points $x_i$ in $\sigma$ that are at a distance strictly less than $r^{\varsigma_n}_{k\mbox{\tiny -NN}}(x)$ to $x$ are in $N_k^{\sigma}(x)$. 
\end{enumerate}

The $k$ nearest neighbour mapping $N^{\sigma}_k(x)$ (which we will sometimes shorten to $N_k(x)$) can be deterministic or stochastic, in which case it will depend on an additional random variable, independent of the sample path. An example of the kind would be to give the sample $\sigma$ a random order, under a uniform distribution on the group of $n$-permutations, and break the distance ties by selecting among the tied neighbours on the sphere the smallest ones under the order selected.

Here is a formal definition of the $k$-NN learning rule:
\begin{eqnarray*}
{g}^{k\mbox{\tiny -NN}}_n(\sigma)(x) &=&
\theta\left[\frac 1k\sum_{x_i\in N_k^{\sigma}(x)}y_i-\frac 12 \right].
\end{eqnarray*}
Above, $\theta$ is the Heaviside function:
\[\theta(t) =\begin{cases} 1,&\mbox{ if }t\geq 0,\\
0,&\mbox{ if }t<0.\end{cases}\]

The $k$-NN rule was historically the first classification learning rule in a standard Borel space whose universal consistency was established, by Charles J. Stone \cite{stone:77}.

\begin{thrm}[C.J. Stone, 1977] The $k$-nearest neighbour classifier is universally consistent in the finite-dimensional Euclidean space whenever $n\to\infty$, $k=k_n\to\infty$, $k/n\to 0$.
\end{thrm}

The $k$-NN classifier is no longer universally consistent in more general separable metric spaces, in fact already in the infinite-dimensional Hilbert space $\ell^2$, as noted in \cite{CG}. An example of this kind (constructed for the needs of real analysis) belongs to Preiss \cite{preiss79}. (See this example adapted for the $k$-NN classifier in \cite{CKP}, sect. 2.) This brings up the question of characterizing those metric spaces in which the $k$-NN classifier is universally consistent, and so far the problem remains open. 

\subsection{Strong consistency\label{strongconsistency}}

Under the --- possibly the most natural --- randomized method of tie-breaking, the $k$-NN classifier is {\em never} strongly universally consistent. Let $(Z_n)$ be a sequence of i.i.d. random variables distributed uniformly in the unit interval $\I=[0,1]$, and independent on data. In case of distance ties, we choose among the points $x_{n_1},x_{n_2},\ldots,x_{n_m}$ at an equal distance to $x$ those points whose corresponding instances $z_{n_i}$ are the smallest. (See for example \cite{CG}, bottom of p. 341.)

\begin{prpstn}
If a sequence of values of $k$, $(k_n)$, goes to infinity sufficiently slowly, then the $k$-NN classifier, under the uniform random tie-breaking using the auxiliary variables $Z_i\in\I$ as above, is not strongly universally consistent in any metric space.
\label{p:notstronglyconsistent}
\end{prpstn}

\begin{proof}
Let the underlying probability measure $\mu$ on $\Omega$ be a Dirac measure concentrated in one point, and let the regression function $\eta$ take a value $p\in (0,1)$, $p\neq 1/2$ at the unique point of the measure support. This way, the nature of the metric space becomes totally irrelevant, as everything reduces to a trivial one-point domain, $\Omega=\{\ast\}$. A sample path in this context is just a Bernoulli sequence $(Y_n)$ of random labels $0$ and $1$ with probability of success $p$, together with an i.i.d. sequence $Z_n\in \I$ of tie-breaking values, the two sequences being independent. The Bayes error for our problem equals $\min\{p,1-p\}$. It is achieved at the Bayes (optimal) classifier, returning the label $1$ if $p>1/2$ and the label $0$ if $p<1/2$. (Here we need the assumption $p\neq 1/2$: for $p=1/2$ any prediction would achieve the Bayes error $1/2$.) Strong universal consistency would require that for a.e. Bernoulli sequence $(Y_n)$ and a.e. tie-breaking sequence $(Z_n)$, the $k$-NN classifier always predicts the Bayes label, starting with some $i$ large enough.

Fix a summable sequence $(\delta_i)$, $\delta_i\in (0,1)$. Choose recursively  sequences $n_i\uparrow\infty$ and $\e_i\downarrow 0$ in such a way that for every $i$, if we randomly choose $n_i$ i.i.d. uniform elements of the interval, $Z_1,Z_2,\ldots,Z_{n_i}$, then with confidence $>1-\delta_i$
\begin{enumerate}
\item at least $\lceil\ln i\rceil$ elements $Z_i$ belong to the interval $[0,\e_i)$, while
\item none of $Z_i$ belong to $[0,\e_{i+1})$.
\end{enumerate}
Now define for each $n$
\[k_n = \lceil \ln i\rceil,\mbox{ if } n_i\leq n <n_{i+1}.\]
The first Borel--Cantelli lemma implies that almost surely, for some $j$ occurs  the event $A_j$ ``a sample path $(Z_i)$ satisfies the conditions (1) and (2) for all $i\geq j$.'' 

Denote $\Theta$ the event ``the $k$-NN classifier returns a wrong label infinitely often''. We will show that at least for some values of $p$, it is an almost sure event.
We will condition on the tie-breaking path $(Z_i)$. Almost surely, $(Z_i)$ is in $A_j$ for some $j$. So let us fix $j$ and a path $(z_i)$ belonging to $A_j$. The properties of $A_j$ imply that, for all $i,m$, such that $j\leq i < m$, the $k=\lceil \ln i\rceil$ smallest elements among $z_1,z_2,\ldots,z_{n_i}$ belong to the interval $(\e_{i+1},\e_i)$, while the $k=\lceil \ln m\rceil$ smallest elements among $z_1,z_2,\ldots,z_i,\ldots,z_{n_m}$ belong to the interval $(0,\e_m)$. Since $\e_m\leq \e_{i+1}$, the two intervals are disjoint, so the sets of tie-breaking values at the steps $n_i$ and $n_m$ are disjoint too, and the subsamples $N^{\sigma_{n_i}}_{k_{n_i}}(\ast)$ and $N^{\sigma_{n_m}}_{k_{n_m}}(\ast)$ of nearest neighbours selected by the classifier to make a prediction at the steps $i$ and $m$ are disjoint (are indexed with disjoints sets of integers). We conclude: the sets of labels of the $k$-nearest neighbours chosen at the moments $n_i$, $i\geq j$ according to our procedure form a sequence of independent random variables with values in $\{0,1\}^{k_{n_i}}$. Consequently, the predictions made at the steps $n_i$, $i\geq j$ also form an independent sequence.

Denote $W_i$ the event ``the $k$-NN classifier returns the wrong label at the step $n_i$ when using the sequence $(z_i)$ for tie-breaking''. According to the above, the sequence of events $(W_i)$, $i\geq j$ is independent. 
The probability for the $k$-NN classifier to return the wrong label (that is, $1$ if $p<1/2$ and $0$ if $p>1/2$) at the step $n_i$, $i\geq j$ is at least $\min\{p,1-p\}^{k_{n_i}}= \min\{p,1-p\}^{\lceil \ln i\rceil}$ (this is the probability of the event where all $k$ nearest neighbours have the same label opposite to the Bayes one). 

Now let $p=e^{-1}\approx 0.368\ldots$. We have
\begin{align*}
\sum_{i=j}^{\infty}\min\{p,1-p\}^{\lceil \ln i\rceil} &= \sum e^{-\lceil \ln i\rceil} \\
&\geq e^{-1}\sum e^{-\ln i} \\
&=e^{-1} \sum_{i=j}^{\infty} \frac 1i,
\end{align*}
a divergent sequence. The events $(W_i)$, $i\geq j$ are independent, the sequence $p(W_i)$ is divergent. The second Borel-Cantelli lemma implies that, almost surely, $W_i$ occur infinitely often. In other words, if $p=e^{-1}$ and our $(z_i)$ is used for tie-breaking, the $k$-NN rule will return the wrong label infinitely often for almost all labelling sequences $(Y_i)$. Since the sequences $(Z_i)$ and $(Y_i)$ are mutually independent, we conclude by the Fubini theorem that our event $\Theta$ occurs with probability one (the same holds in fact whenever $p$ belongs to $[e^{-1},1/2)\cup (1/2,1-e^{-1}]$). 
\end{proof}

In view of this observation, one way to get strong consistency results is to make $k$ grow fast enough. For some results obtained in this direction, see \cite{devroye82}. We do not touch upon this approach in our paper.

Another possibility is to assume that there are no distance ties, that is, there are no atoms and all the spheres have measure zero. This happens in the Euclidean case, for instance, if the underlying distribution has Lebesgue density. Under this assumption, strong consistency for the $k$-NN classifier in the Euclidean space is a result due to Devroye and Gy\"orfi \cite{DG} and to Zhao \cite{Z}. We will extend the same conclusion to all sigma-finite dimensional metric spaces in the sense of Nagata in Section \ref{s:noties}.

Finally, a modified randomized tie-breaking approach to the $k$-NN classifier was proposed by Devroye, Gy\"{o}rfi, Krzy\.{z}ak, and Lugosi in \cite{DGKL}. As before, the data path is enlarged by adding an independent i.i.d. sequence of tie-breaking variables $(Z_n)$ taking value in $\I$. The difference with the previous approach is that the test data point is also modelled not by a single random variable $X\sim\mu$ but a pair of random variables, $(X,Z)$, where $Z$ is independent of $X$ and of the data and follows the uniform distribution on $\I$. In the case of distance ties, the points $X_i$, $i\in J$ all at the same distance from $X$ are ordered in accordance with the corresponding values of $Z_i$, $i\in J$, the closest ones to $Z$ being chosen first. (The previously described approach corresponds to the case of $Z$ taking a constant value zero.)

Under this mode of tie-breaking, the classifier is being built not in $\Omega$ proper but rather in the extended domain $\Omega\times\I$, equipped with the product of $\mu$ and the uniform measure $\lambda$, and whose regression function is the composition of $\eta$ with the projection on the first coordinate. In the Euclidean case $\Omega=\R^d$ it was shown by Devroye {\em et al.} \cite{DGKL} that the resulting classifier, which is, strictly speaking, not the $k$-NN classifier but a modification thereof, converges along almost every sample path to the Bayes classifier on $\Omega\times\I$, obtained by composing the Bayes classifier for $\Omega$ with the first coordinate projection. Even if for any fixed value $Z=z$ the same argument as in our Proposition \ref{p:notstronglyconsistent} shows that the wrong predictions may occur infinitely often, the expected error averaged over $Z\in\I$ converges to zero for almost all sample paths. Thus, if one now wants to obtain a strongly consistent learning rule on $\Omega$ proper, one has to average the predictions along every fibre $\{x\}\times\I$, that is, take the majority vote over all values of the auxiliary variable $Z$. In this approach, essentially, one combines the $k$-NN with ensemble learning.

In Section \ref{zero}, we will establish strong consistency within the above approach for non-Archimedian metric spaces, and the proof shows interesting geometric differences from the Euclidean case.

\section{\label{s:degroot}
Dimension in the sense of de Groot and the Heisenberg group}
The aim of this section is to observe that a complete separable metric space in which the $k$-NN classifier is universally consistent need not be sigma-finite dimensional in the sense of Nagata.
We begin by reminding the important result by C\'erou and Guyader.

\begin{thrm}[C\'erou and Guyader, \cite{CG}] 
Let $\Omega$ be a separable complete metric space equipped with a probability measure $\mu$ (the distribution law of data) and a regression function $\eta\colon\Omega\to [0,1]$ (the conditional probability for a point to be labelled $1$). Suppose further that the regression function 
satisfies the weak Lebesgue--Besicovitch differentiation property:
\begin{equation}
\frac{1}{\mu(B(x,r))}\int_{B(x,r)} \eta(x)\,d\mu(x) \to \eta(x),
\label{eq:lb}
\end{equation}
where the convergence is in measure, that is, for each $\e>0$,
\[\mu\left\{x\in\Omega\colon 
\left\vert\frac{1}{\mu(B(x,r))}\int_{B(x,r)} \eta(x)\,d\mu(x) - \eta(x)
\right\vert >\e\right\}\to 0\mbox{ when } r\downarrow 0.\]
Then the $k$-NN classifier is (weakly) consistent for the supervised learning problem $(\mu,\eta)$ in $\Omega$.
\label{th:CG}
\end{thrm}

Now, some necessary concepts and results related to the Nagata dimension. (For a more detailed presentation with many examples, see Part I of our work \cite{CKP}.)
The following definition is Preiss' generalization \cite{preiss83} of Nagata's original concept. Recall that a family $\gamma$ of subsets of a set $\Omega$ has {\em multiplicity} $\leq n$ if every point of $\Omega$ is contained in at most $n$ elements of $\gamma$.

\begin{dfntn} Let $\Omega$ be a metric space and $X$ a metric subspace, let $\delta\in\N$ and $s>0$. Then $X$ has {\em Nagata dimension $\leq\delta$ on the scale $s$ inside of} $\Omega$ if every finite family of closed balls in $\Omega$ with centres in $X$ and radii $<s$ admits a subfamily having multiplicity $\leq\delta+1$ in $\Omega$ which covers all the centres of the original balls. The Nagata dimension of $X$ within $\Omega$ on the scale $s>0$, denoted $\dim^s_{Nag}(X,\Omega)$ or sometimes simply $\dim_{Nag}(X,\Omega)$, is the smallest $\delta$ such that $X$ has Nagata dimension $\leq\delta$ on the scale $s$ inside $\Omega$. We say that a subspace $X$ has a finite Nagata dimension in $\Omega$ if $X$ has finite dimension in $\Omega$ on some suitable scale $s>0$. 
\label{d:dimnagata}
\end{dfntn}

Here is a reformulation that we will use. A family of balls in a metric space is {\em disconnected} if the centre of each ball of the family does not belong to any other ball. 

\begin{prpstn}
For a subspace $X$ of a metric space $\Omega$, one has
\[\dim^s_{Nag}(X,\Omega) \leq\delta\]
if and only if every disconnected family of closed balls in $\Omega$ of radii $<s$ with centres in $X$ has multiplicity $\leq\delta+1$.
\label{ex:famdesconexa}
\end{prpstn}

For a proof, see e.g. \cite{CKP}, Prop. 7.2. 
Here is another important property: the Nagata dimension does not increase when we form the closure of a subspace.

\begin{prpstn}[See \cite{CKP}, Prop. 7.4]
Let $X$ be a subspace of a metric space $\Omega$, satisfying $\dim^s_{Nag}(X,\Omega)\leq \delta$. Then $\dim^s_{Nag}(\bar X,\Omega)\leq \delta$, where $\bar X$ is the closure of $X$ in $\Omega$.
\label{p:closurenagata}
\end{prpstn}

\begin{dfntn}[Preiss, \cite{preiss83}]
A metric space $\Omega$ is said to be {\em sigma-finite dimensional in the sense of Nagata} if $\Omega=\cup_{i=1}^{\infty}X_n$, where every subspace $X_n$ has finite Nagata dimension in $\Omega$ on some scale $s_n>0$ (where the scales $s_n$ are possibly all different).
\end{dfntn}

\begin{rmrk}
Because of Proposition \ref{p:closurenagata}, we can assume all $X_n$ to be closed. Also, it is easy to see that the union of two subspaces having finite Nagata dimension each also has a finite Nagata dimension (Prop. 7.5 in \cite{CKP}), so we can in addition assume that $X_n$ form an increasing chain.
\label{r:open}
\end{rmrk}

\begin{rmrk}
In view of the preceding remark, the Baire Category argument implies that every complete metric space $\Omega$ that is sigma-finite dimensional in the sense of Nagata contains a non-empty open subspace that has finite Nagata dimension in $\Omega$.
\label{r:baire}
\end{rmrk}

Now we can remind the theorem of Preiss.

\begin{thrm}[Preiss \cite{preiss83}] Let $\Omega$ be a complete separable metric space. Then the following two properties are equivalent.
\begin{enumerate}
\item
For every locally finite Borel measure $\mu$ on $\Omega$, every $L^1(\mu)$-function $f\colon\Omega\to\R$ satisies the strong Lebesgue--Besicovitch differentiation property: for $\mu$-a.e. $x\in\Omega$,
\begin{equation}
\frac{1}{\mu(B(x,r))}\int_{B(x,r)} f(x)\,d\mu(x) \to f(x)\mbox{ as }r\downarrow 0.
\label{eq:lbs}
\end{equation}
\item $\Omega$ is sigma-finite dimensional in the sense of Nagata.
\end{enumerate}
\label{th:preiss}
\end{thrm}

It should be noted that the original note of Preiss \cite{preiss83} only contained a brief sketch of the proof of the implication (1)$\Rightarrow$(2). The implication (2)$\Rightarrow$(1) was worked out in detail by Assouad and Quentin de Gromard in \cite{AQ} for the case of finite Nagata dimension (from this, the deduction of the sigma-finite dimensional case is straightforward). 

By combining theorems \ref{th:preiss} and \ref{th:CG}, one obtains:

\begin{crllr}
The $k$-nearest neighbour classifier is (weakly) universally consistent in every complete separable metric space sigma-finite dimensional in the sense of Nagata.
\label{c:preiss-cerou}
\end{crllr}

In Part I \cite{CKP} we have given a direct proof of this result along the geometric ideas of the original proof of Stone \cite{stone:77}.

Note that Preiss' result asserts a strong version of the Lebesgue--Besicovitch property, while the result of C\'erou and Guyader only requires the weak version of it as an assumption. Turns out, there is a class of metric spaces that ``fills the gap'' between the two. For that, we need to give some more definitions.

\begin{dfntn}[\cite{dG}; \cite{AQ}, 3.5]
Let $\delta\in\N$.
A metric space $\Omega$ has de Groot dimension $\leq\delta$ if it satisfies the following property. For every closed ball $\bar B(a,r)$ in $\Omega$ with centre $a$ and radius $r>0$, if $x_1,\ldots,x_{\delta+1}\in \bar B(a,r)$, then there are $i\neq j$ with $d(x_i,x_j)\leq r$.
\end{dfntn}

\begin{prpstn}[Prop. 3.1 in \cite{AQ}] A metric space $\Omega$ has de Groot dimension $\leq\delta$ if and only if every finite family of closed balls having the same radii admits a subfamily covering all the centres of the original balls and having multiplicity $\leq\delta+1$.
\end{prpstn}

\begin{proof} Necessity: let $\bar B(x_1,r),\ldots,\bar B(x_N,r)$ be a finite family of closed balls having the same radius. Take any maximal disconnected subfamily of those balls. It covers all the centres by maximality (here we use the fact that the radii of all the balls are the same). Also, this maximal disconnected subfamily has multiplicity $\leq\delta+1$ because of our assumption on de Groot dimension: assuming there were $x$ belonging to $\delta+2$ balls, the $r$-ball centred at $x$ of radius $r$ would contain $\delta+2$ points two by two at a distance $>r$ from each other.

Sufficiency: apply the property to the family of balls $\bar B(x_i,r)$, $i=1,2,\ldots,\delta+1$, where $x_i\in \bar B(x,r)$. All of the above closed balls contain $x$, so at least one of those balls, say $\bar B(x_i,r)$, will be missing from a subfamily containing all the centres; then $x_i\in\bar B(x_j,r)$, $j\neq i$, so $d(x_i,x_j)\leq r$.
\end{proof}

Thus, in view of Proposition \ref{ex:famdesconexa}, de Groot dimension of a metric space is always bounded by the Nagata dimension on the scale $+\infty$. For the space $\R^n$ equipped with an arbitrary norm, the two dimensions are equal (\cite{AQ}, 4.9).
In a more general case, in fact, already in the infinite-dimensional Hilbert space $\ell^2$, the distinguishing examples are easy to construct. 

\begin{xmpl}
The convergent sequence $2^{-n}e_n$, $n\geq 0$, where $e_n$ are elements of the standard orthonormal basis in the Hilbert space $\ell^2$, together with the limit $0$, equipped with the induced metric, has infinite Nagata dimension on every scale $s>0$. Indeed, each closed ball of radius $2^{-n}$, centred at $2^{-n}e_n$, contains $0$ as the only other element of the space, and so admits no subfamily of finite multiplicity containing all the centres. 

At the same time, this sequence has de Groot dimension $2$. Call $n$ the {\em index} of a point $x=2^{-n}e_n$, and let the index of zero be infinite. Denote the index $i(x)$.
Given a closed ball of centre $a$ in this space and three points inside the ball, order them according to the increasing index, $x_1,x_2,x_3$. If now $i(a)\leq i(x_1)$, then $x_2$ and $x_3$ are closer to each other than $x_3$ is to $a$. And if $i(x_1)<i(a)$, then the distance between $x_2$ and $x_3$ is smaller than between $a$ and $x_1$. (And notice that de Groot dimension is not equal to one as the example of a ball of radius $1/2$ centred at $x=2^{-3}e_3$ and containing two points, $x_1=2^{-1}e_1$ and $x_2=2^{-2}e_2$ shows.)

This space is complete (even compact) and sigma-finite dimensional in the sense of Nagata being the union of countably many singletons: a singleton trivially has Nagata dimension zero in every ambient metric space.
\end{xmpl}

A source of metric spaces of finite de Groot dimension is provided by the doubling metric spaces. 

\begin{dfntn}
A metric space $X$ is {\em doubling} if there is a constant $C>0$ such that for every $x\in X$ and $r>0$, the closed ball $\bar B(x,r)$ can be covered with at most $C$ closed balls of radius $r/2$.
\label{def:C}
\end{dfntn}

The following is a simple exercise. (Cover a closed $r$-ball with $\leq C$ many $r/2$-balls and notice that among any $C+1$ points, at least two belong to the same closed $r/2$-ball.)

\begin{prpstn}
Every doubling metric space has finite de Groot dimension (bounded by the constant $C$ from definition \ref{def:C}).
\end{prpstn}

Metric spaces of finite de Groot dimension satisfy the weak Lebesgue-Besicovitch differentiation property.

\begin{thrm}[Assouad and Quentin de Gromard, \cite{AQ}, Prop. 3.3.1(b)+Prop. 3.1]
Let a complete separable metric space $\Omega$ have finite de Groot dimension. Then for every probability Borel measure $\mu$ on $\Omega$, every $L^1(\mu)$-function $f\colon\Omega\to\R$ satisies the weak Lebesgue--Besicovitch differentiation property: 
\begin{equation}
\frac{1}{\mu(B(x,r))}\int_{B(x,r)} f(x)\,d\mu(x) \to f(x)
\label{eq:lbg}
\end{equation}
in measure, when $r\downarrow 0$.
\label{th:assouad}
\end{thrm}

Combining this result with that of C\'erou and Guyader (Theorem \ref{th:CG} above), we arrive at:

\begin{crllr}
The $k$-nearest neighbour classifier is universally consistent in every complete separable metric space having finite de Groot dimension.
\label{c:degrootkNN}
\end{crllr}

It would be certainly interesting to give a direct proof of the result in the spirit of Stone. Moreover, the versions of de Groot dimension on a given scale and of sigma-finite dimensional spaces in the sense of de Groot that exactly parallel the definition of Preiss can be easily stated, so it is natural to ask a number of questions about such spaces. For instance, is it true that a metric space has the weak Lebesgue--Besicovitch property if and only if it is sigma-finite dimensional in the sense of de Groot? See the concluding Section \ref{s:conjecture} for an exact formulation.

An example of a complete separable metric space of finite de Groot dimension that is not sigma-finite dimensional in the sense of Nagata is provided by the Heisenberg group $\Hg$ equipped with one of the natural metrics that we now proceed to describe.

Topologically, the Heisenberg group $\Hg$ is identified with the Euclidean space $\R^{3}$, and is equipped with the following group multiplication:
\begin{equation}
(x,y,z)\cdot (x^{\prime},y^{\prime},z^{\prime}) = 
\left(x+x^{\prime},y+y^{\prime},z+z^{\prime}+C xy^{\prime}-C yx^{\prime}\right).
\label{eq:multi1}
\end{equation}
Here $x,x^{\prime},y,y^{\prime},z,z^{\prime}\in\R$, and $C\neq 0$ is a real constant. Different choices of $C$ result in algebraically isomorphic groups: a group isomorphism from the above version to the one determined by the constant $C^\prime$ is given by a linear map multiplying each vector by $C/C^\prime$.

The operation in (\ref{eq:multi1}) clearly makes $\Hg$ into a topological group, in fact a Lie group, when equipped with the Euclidean topology.

For all values of $C$ with $\abs C\leq 4$ the formula
\[\left\vert (x,y,z) \right\vert_{\Hg} = \left((x^2+y^2)^2+z^2 \right)^{1/4}\]
defines a {\em group norm} on $\Hg$, in the sense that $\abs{p^{-1}}_{\Hg}=\abs{p}_{\Hg}$ and
\[\abs{p\cdot q}_{\Hg}\leq \abs{p}_{\Hg}+\abs{q}_{\Hg}.\]
The latter is a consequence of the following particular case of a result of Cygan \cite{cygan} (using notation and concepts from, and better be looked at jointly with, the article \cite{K}):
\[\left(\left[(x+x^{\prime})^2+(y+y^{\prime})^2\right]^2+16 (z+z^{\prime}+ xy^{\prime}-yx^{\prime})^2\right)^{1/4}\leq 
\left((x^2+y^2)^2+16 z^2 \right)^{1/4}+\left((x^{\prime 2}+y^{\prime 2})^2+16 z^{\prime 2} \right)^{1/4}.
 \]
Given an expression on the right of Eq. (\ref{eq:multi1}), denote $\ve=\pm 1$ the product of the signs of $z+z^\prime$ and of $C xy^{\prime}-C yx^{\prime}$. Assuming $\abs C\leq 4$, the norm of the product $(x,y,z)\cdot (x^{\prime},y^{\prime},z^{\prime})$ is less than or equal to
\[\left[\left[(x+x^{\prime})^2+(y+y^{\prime})^2\right]^2+16 \left(\frac{\ve z}{4}+\frac{\ve z^{\prime}}{4}+ xy^{\prime}-yx^{\prime}\right)^2\right]^{1/4},\]
and applying Cygan's inequality, we arrive at the product of norms of $(x,y,z)$ and $(x^{\prime},y^{\prime},z^{\prime})$.

Consequently, a left-invariant metric on $\Hg$ is defined by
\[d(p,q)=\abs{p^{-1}\cdot q}_{\Hg},\]
and is clearly compatible with the Euclidean topology. This distance is known as a ({\em Cygan}-){\em Kor\'anyi} distance. Thus, it is the unique left-invariant metric such that
\[d(e,p)=\abs{p}_{\Hg}.\]

It is well-known and readily seen that the group $\Hg$ equipped with a Cygan--Kor\'anyi distance is doubling. In fact, the doubling property holds for any compatible left-invariant metric on $\Hg$ that is {\em homogeneous} in the sense that if we apply to the group the transformation $(x,y,z)\mapsto (tx,ty,t^2z)$ for $t>0$, then the distance between any pair of points increases by the factor of $t$. (It can actually be shown that every such metric is automatically compatible with the Euclidean topology, see \cite{leD}.) In this form, the doubling property is enough to establish for a single ball of radius $r=1$ say centred at zero, and it follows from local compactness of the Euclidean space. As the Cygan--Kor\'anyi metric is both left-invariant and homogeneous (an easy calculation), the statement follows. In particular, we conclude from the result of Assouad and Quentin de Gromard (Theorem \ref{th:assouad}):

\begin{crllr}
The Heisenberg group $\Hg$ equipped with a Cygan--Kor\'anyi metric satisfies the weak Lebesgue--Besicovitch property for every Borel probability measure $\mu$ and every $L^1(\mu)$-function.
\label{c:HLB}
\end{crllr}

According to the result of C\'erou and Guyader (Theorem \ref{th:CG}), we now have:

\begin{crllr}
The $k$-NN learning rule is universally consistent in the Heisenberg group $\Hg$ equipped with a Cygan--Kor\'anyi metric.
\end{crllr}

At the same time, the metric space $\Hg$ with a Cygan--Kor\'anyi distance need not be sigma-finite dimensional in the sense of Nagata. 
For the next result, we choose a version of the group law corresponding to the value $C=-2$ in the multiplication formula (\ref{eq:multi1}), following \cite{KR}. Thus, 
\begin{equation}
(x,y,z)\cdot (x^{\prime},y^{\prime},z^{\prime}) = 
\left(x+x^{\prime},y+y^{\prime},z+z^{\prime}-2 xy^{\prime}+2 yx^{\prime}\right).
\label{eq:multi}
\end{equation}
Essentially, by fixing $C$, we select a version of the Cygan--Kor\'anyi metric, because the groups are all isomorphic between themselves for different values of $C\neq 0$.

\begin{lmm}[Kor\'anyi and Reimann, \cite{KR}, p. 17; Sawyer and Wheeden, \cite{SW}, Lemma 4.4, p. 863] 
Let $C=-2$.
There exists a sequence $(p_n)$ of elements of $\Hg$ with $r_n=\abs{p_n}_{\Hg}\to 0$ so that the family of balls $\bar B(p_n,r_n)$ is disconnected. 
\label{th:koranyi}
\end{lmm}

We find it useful to present a proof, following \cite{KR} and somewhat expanding the argument.

\begin{proof}
By identifying $\R^2$ with the complex plane $\C$, we can write the multiplication law (\ref{eq:multi}) in the group $\Hg=\C\times\R$ as
\[(z,t)(z^\prime,t^\prime)=(z+z^\prime,t+t^\prime + 2\,\mathrm{Im}\, z \bar z^\prime).\]
The neutral element of the group is $(0,0)$, and the inverse of $(z,t)$ is simply $(-z,-t)$. Consequently,
 the formula for the left-invariant Cygan--Kor\'anyi metric becomes:
\begin{align*}
d((z,t),(z^\prime,t^\prime)) &= \left\vert (z,t)^{-1}(z^\prime,t^\prime) \right\vert_{\Hg} \\
&= \left\vert (-z,-t)(z^\prime,t^\prime) \right\vert_{\Hg} \\
&= \left\vert (-z+z^\prime,-t+t^\prime-2\,\mathrm{Im}\, z \bar z^\prime) \right\vert_{\Hg} \\
&=\left(\left\vert-z+z^\prime \right\vert^4 +\left\vert -t+t^\prime -2\,\mathrm{Im}\,z \bar z^\prime\right\vert^2\right)^{1/4}.
\end{align*}

Let $(z,t)$ and $(z^\prime,t^\prime)$ be two points on the unit sphere of $\Hg$ around the neutral element $0$ that are different from $(0,0,\pm 1)$ (so that $z,z^\prime\neq 0$). Notice that $\mathrm{Re} \,z \bar z^\prime$ is the inner product of $z$ and $z^\prime$  as vectors of $\R^2$. 
As $r\downarrow 0$, we have up to the second order terms in $r$:
\begin{align*}
d((rz,r^2t),(z^\prime,t^\prime))^4 &= \left\vert -rz + z^\prime\right\vert^4 +
\left\vert -r^2 t + t^\prime -2\, \mathrm{Im}\,rz\bar z^\prime \right\vert^2 \\
&= \left ( r^2 \abs{z}^2 + \abs{z^\prime}^2 - 2 r \mathrm{Re}\, z \bar z^\prime \right )^2 + \left( -r^2 t + t^\prime -2\, \mathrm{Im}\,rz \bar z^\prime \right)^2\\
& \overset{O(r^2)}{\approx}
\abs{z^\prime}^4 - 4\abs{z^\prime}^2 r \mathrm{Re}\,z \bar z^\prime
+ t^{\prime 2} - 4t^\prime r\mathrm{Im}\,z\bar z^\prime \\
& = 1 - 4r \left(\abs{z^\prime}^2\mathrm{Re}\,z \bar z^\prime +t^\prime\mathrm{Im}\,z\bar z^\prime\right).
\end{align*}
If the bracketed term on the right is strictly negative,
\begin{equation}
\abs{z^\prime}^2\mathrm{Re}\,z \bar z^\prime +t^\prime\mathrm{Im}\,z\bar z^\prime < 0,
\label{eq:condition}
\end{equation}
 then for sufficiently small $r>0$
\[d((rz,r^2t),(z^\prime,t^\prime))>1,\]
so for any $\rho>0$, using the homogeneity property of the metric,
\begin{equation}
d((r\rho z,r^2\rho^2 t),(\rho z^\prime,\rho^2t^\prime))>\rho.
\label{eq:rho}
\end{equation}
Since the complex number $t^\prime+\abs{z^\prime}^2 i$ has modulus one, it can be written as $e^{\psi i}$, so the condition in Eq. (\ref{eq:condition}) becomes
\begin{equation}
\mathrm{Im}(e^{\psi i}z\bar z^{\prime})<0.
\label{eq:condition2}
\end{equation}
Now we define two sequences of reals
\[\psi_j=-\frac{\pi}2\frac{1}{(j+1)^2}+\pi,~~\theta_j=\frac{\pi}2\frac{j-1}j
\]
and a sequence of points on the unit sphere in $\Hg$ 
\[(z_j,t_j)=\left(e^{\theta_j i}\sqrt{\sin \psi_j}, \cos\psi_j\right).
\]
Notice that
\[e^{\psi_ji}=t_j+\abs{z_j}^2 i.
\]
Since for $n>j$ we have
\[\pi < \theta_{j+1}-\theta_j+\psi_j\leq \theta_n-\theta_j+\psi_j<\frac{3\pi}2,
\]
(there is a small typo in the second displayed formula on p. 18 in \cite{KR}),
it follows that
\[\mathrm{Im}(e^{\psi_ji}z_n\bar z_j)\leq \mathrm{Im}(e^{\psi_ji}z_{j+1}\bar z_j)<0.\]
Now the radii $r_j>0$ are being chosen recursively, using Eq. (\ref{eq:rho}), in such a way that each element $(r_jz_j,r_j^2t_j)$ is outside the finitely many closed balls already selected.
\end{proof}

Since all of the above closed balls $\bar B(p_n,r_n)$ contain zero (the identity of $\Hg$), the Nagata dimension of $\Hg$ is infinite by Prop. \ref{ex:famdesconexa}, as was noted by Assouad and Quentin de Gromard \cite{AQ}, 4.7(f).
But in fact, the construction implies more.

\begin{crllr}
The group $\Hg$ equipped with the Cygan--Kor\'anyi metric is not sigma-finite dimensional in the sense of Nagata.
\label{c:notsigmafd}
\end{crllr}

\begin{proof}
Assuming $\Hg$ were sigma-finite dimensional, by our Remark \ref{r:baire}, it would contain a non-empty open subset $U$ which has finite Nagata dimension in $\Hg$. Select any $p\in U$. Since the metric is left-invariant and so the left translation $q\mapsto p^{-1}\cdot q$ is an isometry, the set $p^{-1}\cdot U$ also has finite Nagata dimension. Since this set is a neighbourhood of identity, it contains all elements of the sequence $(x_n)$ chosen as in Theorem \ref{th:koranyi}, beginning with $n$ large enough. This contradicts the finite dimensionality of the set $p^{-1}\cdot U$ inside $\Hg$ in the sense of Nagata.
\end{proof}

Thus, the Heisenberg group $\Hg$ provides an example of a metric space possessing the weak Lebesgue--Besicovitch property --- in particular, on which the $k$-NN classifier is universally (weakly) consistent --- and which is not sigma-finite dimensional.

\begin{rmrk}
The influential 1983 paper by Preiss \cite{preiss83} mentioned that it was unknown whether a complete separable metric space $\Omega$ satisfies the weak Lebesgue--Besicovitch differentiation property for every Borel locally finite measure if and only if $\Omega$ satisfies the strong Lebesgue--Besicovitch differentiation property for every Borel locally finite measure. The later developments have shown the answer to be negative, in fact the Heisenberg group with the Cygan--Kor\'anyi metric provides a distinguishing example in view of Corollary \ref{c:HLB}, Corollary \ref{c:notsigmafd} and Preiss's Theorem \ref{th:preiss}, (1)$\Rightarrow$(2). This fact must be well known to the specialists, even if we have not found it mentioned explicitly anywhere. 
\end{rmrk}

\section{\label{s:noties}Strong consistency in the absence of distance ties}

A probability measure $\mu$ on a metric space $\Omega$ has a zero probability of distance ties if the measure of every sphere $S_r(x)$, $x\in \Omega$, $r\geq 0$ is zero. In particular, such a measure is non-atomic (the case $r=0$).
In this section, we will show that the result by Devroye and Gy\"orfi \cite{DG} and Zhao \cite{Z} about the strong universal consistency of the $k$-NN classifier in the Euclidean space in the absence of distance ties is valid in all complete separable sigma-finite dimensional metric spaces in the sense of Nagata -- again, in the case where distance ties occur with zero probability. We will follow the presentation of the proof of Theorem 11.1 in \cite{DGL}, however, as to be expected, the extension requires certain technical modifications, not all of which concern Lemma \ref{lem:stone_lemma_strong} below.

\begin{thrm}
Under the zero probability of distance ties, the $k$-NN learning rule is strongly universally consistent in every complete separable metric space that is sigma-finite dimensional in the sense of Nagata.
\label{th:strongsigma}
\end{thrm}

\begin{rmrk}
The result is certainly of interest in the setting of all finite-dimensional normed spaces (not just the Euclidean ones), because in such a space there are no distance ties whenever the underlying distribution has density with regard to the Lebesgue measure. It is hard to think of a similar natural condition for sigma-finite dimensional metric spaces beyond the normed spaces case. One of the most interesting classes -- and in which the distance-based classifiers are of practical interest \cite{martinez} -- is given by the non-Archimedian metric spaces, satisfying the strong triangle inequality, which are essentially the metric spaces of Nagata dimension zero. It is not difficult to see that a non-Archimedian metric on a separable space only takes a countable number of distinct values. (Indeed, given such a space, $\Omega$, choose a contable dense subset $X$ and apply the strong triangle inequality to deduce that for any $x,y\in\Omega$ there are $a,b\in X$ with $d(x,y)=d(a,b)$.) This means the distance ties will always occur with strictly positive probability. A rather natural example where the ties are overwhelming was worked out by us in Part I \cite{CKP}, Example 6.4.
\end{rmrk}

Recall from Section \ref{s:rules} that strong consistency of a learning rule $(g_n)$ means that along $\tilde\mu^{\infty}$-almost every infinite labelled sample path $\sigma_{\infty}\in\Omega^{\infty}\times \{0,1\}^{\infty}$, the learning error converges to the Bayes error:
\[\mbox{err}_{\mu,\eta}(g_n(\sigma_n))\to \ell^{\ast}_{\mu,\eta}.\]
Here $\ell^{\ast}_{\mu,\eta}$ is the Bayes error of the learning problem $(\mu,\eta)$, and $\mbox{err}_{\mu,\eta}(g_n(\sigma_n))$ is the error of the classifier given by the learning rule on the sample input $\sigma_n$, the initial $n$-segment of the path $\sigma_{\infty}$. The convergence here is that of a sequence of reals.

Getting back to the $k$-NN learning rule, denote $\eta_n$ the approximation to the regression function:
\[\eta_n(X)=\frac{1}{k}\sum_{i=1}^{n} \mathbb{I}_{\{X_i \in N_k(X)\}}Y_i,\]
where the sum is over all $k$ nearest neighbours of $X$. We have a classical estimate valid in all metric spaces (see \cite{CG}, Proposiion 1.1):
 \begin{align*}
 \mbox{err}_{\mu,\eta}(g_n) - \ell^{\ast}_{\mu,\eta} \leq 2 \mathbb{E}_{\mu}\bigg\{|\eta(X) - \eta_n(X)| \bigg| D_n \bigg\}.
 \end{align*}
Therefore, the strong consistency would follow if we could show that along almost every sample path,
\[\mathbb{E}_{\mu}|\eta(X) - \eta_n(X)|\to 0.\]

A sigma-finite dimensional metric space $\Omega$ can be represented as the union of a countable increasing chain of measurable (even closed should we wish, see Remark \ref{r:open}) subspaces $(F_m)$, each having finite Nagata dimension in $\Omega$, in such a way that $\mu(F_m)\to 1$. Thus, the strong consistency would follow if we could prove that for each fixed $m$, along almost every sample path,
\[\mathbb{E}_{\mu}
\left\{ \left\vert\eta(X) - \eta_n(X)\right\vert X\in F_m\right\}\to 0,\]
where the expectation is conditional, that is, essentially, a normalized integral over $F_m$. The way to prove this is through the Borel--Cantelli lemma: we want to show that the expected value of the difference $\left\vert\eta(X) - \eta_n(X)\right\vert $ over $F_m$ normally concentrates in $n$. We have no control over the rate of convergence of this difference to zero, so it may be very slow, but what matters is that it should be roughly uniform: if for every $\e>0$, starting with $n$ sufficiently large, the probability of a deviation larger than $\e$ is of the order $\exp(-n\e^2)$, we are done: for almost every sample path, beginning with some $n$, the deviation over $F_m$ will be below $\e$. Thus, the following lemma, modelled on Theorem 11.1 in \cite{DGL}, will settle the proof of Theorem \ref{th:strongsigma}, and the rest of the Section will be just devoted to a proof of lemma.

\begin{lmm}
Let $\Omega$ be a complete separable metric space, and let $Q$ be a Borel subset. Suppose $Q$ has Nagata dimension $\leq\beta$ in $\Omega$ on a scale $s$. Let $\mu$ be a probability measure on $\Omega$ with zero probability of ties, and let $\eta\colon \Omega\to [0,1]$ be a regression function. Suppose $\mu(Q)>0$. Let $\tilde\mu$ be a probability measure on $\Omega\times\{0,1\}$ corresponding to $(\mu,\eta)$.
For $\varepsilon > 0$, whenever $k,n \rightarrow \infty$ and $k/n \rightarrow 0$, there is a $n_0$ such that for $n > n_0$,
 \begin{align*}
 \mathbb{P}\bigg( \E_{\tilde\mu}\left\{\left\vert\eta(X) - \eta_n(X)\right\vert X\in Q\right\}
 > \varepsilon \bigg) \leq 4 e^{-\frac{n\varepsilon^2\mu(Q)^2}{18 (\beta+1)^2}}.
 \end{align*}
\label{l:main}
\end{lmm}

Let $\mu$ be a Borel probability measure on a complete separable metric space $\Omega$.
Let $0< \alpha \leq 1$. We define
 \begin{align}
 \label{eqn:r_alpha}
 r_{\alpha}(x) = \inf\{r >0 : \mu(B(x,r)) \geq \alpha \}.
 \end{align}	

 \begin{lmm}
 \label{lem:ties_zero}
 Let $\mu$ be a probability measure with zero probability of ties. Then $\mu(B(x,r_{\alpha}(x))) = \alpha$ for every $x$.
 \end{lmm}
 \begin{proof}
Clearly, $r_{\alpha}(x)>0$.
The measure of every open ball of radius $<r_{\alpha}(x)$ is strictly less than $\alpha$.
By the sigma-additivity of $\mu$, the measure of the open ball of radius $r_{\alpha}(x)$ is $\leq\alpha$, and the measure of the corresponding closed ball $\bar B(x,r_{\alpha}(x))$ is $\geq\alpha$. By our assumption, the sphere is a null set, so the two values are equal.
 \end{proof}

 \begin{lmm}
 \label{lem:r_alpha}
The real-valued function $r_{\alpha}$ defined as in \eqref{eqn:r_alpha} is 1-Lipschitz continuous and converges to zero as $\alpha \rightarrow 0$ at each point of the support of the measure.  
 \end{lmm}

 \begin{proof}
Since $B(y,r_{\alpha}(y))\subseteq B(y,\rho(x,y)+r_{\alpha}(x))$ (the latter ball contains $B(x,r_{\alpha}(x))$ and so has measure $\geq \alpha$), we have $r_{\alpha}(y) \leq \rho(x,y)+ r_{\alpha}(x)$. Therefore, $r_{\alpha}$ is $1$-Lipschitz. The second assertion is clear.
 \end{proof}

The following technical result is an analogue of Lemma 11.1 in \cite{DGL}.

 \begin{lmm}
 \label{lem:stone_lemma_strong}
Let $\Omega$ be a complete separable metric space and let
$Q$ be a Borel subset having Nagata dimension $\leq\beta$ in $\Omega$ on the scale $s$. Assume that $\mu$ is a probability measure on $\Omega$ with zero probability of ties. 
 For $y \in \Omega$, define 
 \begin{align*}
 D(y,\alpha) &= \{ x \in \Omega : y \in B(x,r_{\alpha}(x)) \} \\
&= \{x\in\Omega\colon d(x,y)< r_{\alpha}(x)\}.
 \end{align*}
 Then $\mu(D(y,\alpha)\cap Q) \leq (\beta+1) \alpha$ for all $\alpha$ small enough. 
 \end{lmm}

 \begin{proof}
First of all, notice that the set $D(y,\alpha)$ is open, so it makes sense to talk of its measure. Indeed, if $x\in D(y,\alpha)$, then the open ball of radius $\delta=(1/2)\left[r_{\alpha}(x)-d(x,y)\right]>0$ around $x$ also belongs to $D(y,\alpha)$: every element $x^\prime$ of such a ball satisfies \[r_{\alpha}(x^\prime)>r_{\alpha}(x)-\delta \geq d(x,y)+\delta \geq d(x^\prime,y)\]
(the first inequality is due to Lemma \ref{lem:r_alpha}, the rest follow from the triangle inequality). 

Now let $\varepsilon >0$. By Luzin's theorem, there is a compact set $K \subseteq D(y,a)\cap Q\cap \supp \mu$ such that  $\mu(D(y,a)\cap Q \setminus K) < \varepsilon$. As $\varepsilon>0$ is arbitrary, we need to only get the desired upper bound for $\mu(K)$. 
 
It follows from Lemma \ref{lem:r_alpha} that $r_{\alpha}$ 
converges to 0 uniformly on $K$ when $\alpha$ goes to 0. Choose $\alpha_0 >0$ such that for $0 < \alpha \leq \alpha_0$, we have $r_{\alpha}(x) < s$ for all $x \in K$.
 
Every open ball $B(x,r_{\alpha}(x))$ centered at $x \in K$ contains $y$, therefore 
 \begin{align}
 \label{eqn:subset_r_alpha}
 \bar{B}(x,d(x,y)) \subseteq B(x,r_{\alpha}(x)).
 \end{align}
Let $D = \{ a_n: n \in \mathbb{N}\}$ be a countable dense subset of $K$. Since $Q$ has metric dimension $\beta$ in $\Omega$ on the scale $s$, (\ref{eqn:subset_r_alpha}) implies that for every $n$ there exists a set of $\leq \beta+1$ centers $\{x_1^{n},\ldots,x_{\beta+1}^{n}\} \subseteq \{a_1,\ldots,a_n\}$ such that the closed balls $\bar{B}(x_i^n,d(x_i^n,y))$, $i=1,2,\ldots,\beta+1$ cover 
$\{a_1,\ldots,a_n\}$.

As $K$ is compact, we can recursively select a subset of indices $I\subseteq\N$ so that each sequence of centres $x_i^n$, $i=1,2,\ldots,\beta+1$, $n\in I$ converges to some point $x_i\in K$.
We claim that the union of closed balls $\bar{B}(x_i, d(x_i,y) ), 1 \leq i \leq \beta+1$ covers $K$, which will finish the proof in view of the inclusion (\ref{eqn:subset_r_alpha}). 

As closure of the finite union is the union of closures and since the balls are closed, it is enough to show that $D=\{a_m\}_{m \in \mathbb{N}}$ is contained in the union of $\bar{B}(x_i, d(x_i,y) ), \allowbreak 1 \leq i \leq \beta+1$. Fix $m$. There are $i_0\in \{1,2,\ldots,\beta+1\}$ and an infinite set of indices $J\subseteq I$ such that $a_m$ belongs to all the balls $\bar{B}(x_{i_0}^{n},d(x_{i_0}^{n},y))$, $n\in J$.
It follows that
 \begin{align*}
d(a_m,x_0)&=\lim_{n\in J}d(a_m,x_{i_0}^{n})\\
& \leq \lim_{n\in J}d(x_{i_0}^{n},y)\\ &
=d(x_{i_0},y),
\end{align*}
so $a_m\in \bar{B}(x_{i_0}, d(x_{i_0},y) )$.
 \end{proof}
 
Now, to the proof of Lemma \ref{l:main}. As in Eq. (\ref{eqn:r_alpha}), denote 
$r_{k/n}(x)$ the unique solution to the equation
\[\mu(B(x,r_{k/n}(x))=\frac kn\]
(cf. Lemma \ref{lem:ties_zero}).
Let ${\eta}_{n}^*$  be another approximation of $\eta$,
 \begin{align}
 \label{eqn:eta_star}
 {\eta}_{n}^*(X) = \frac{1}{k}\sum_{i=1}^{n} \mathbb{I}_{\{\rho(X_i,X) < r_{k/n}(X)\}} Y_i.
 \end{align} 
 By the triangle inequality, 
 \begin{align}
 |\eta(X) - \eta_n(X)| \leq |\eta(X) - {\eta}_n^*(X)| + |{\eta}_n^*(X) - \eta_n(X)|. 
\label{eq:triangle}
 \end{align}
Like in Eq. (\ref{eq:rknn}), denote $r_{k\mbox{\tiny -NN}}(x)$ the smallest radius of a closed ball around $x$ containing at least $k$ nearest neighbours of $x$ (we suppress the symbol of the sample). In the absence of distance ties, the closed $r_{k\mbox{\tiny -NN}}(x)$-ball a.s. contains exactly $k$ nearest neighbours. Of the two closed balls around $x$, one of radius $r_{k\mbox{\tiny -NN}}(x)$ and the other of radius $r_{k/n}(x)$, one is necessarily contained in the other, so the symmetric difference, which we tentatively denote $\Delta(x)$, is just the set-theoretic difference of the two balls, though we do not know in which order. With this in mind, we have for the second term on the right-hand side of above equation (\ref{eq:triangle}), 
 \begin{align}
  |{\eta}_n^*(X) - \eta_n(X)| & = \frac{1}{k}\bigg|\sum_{i=1}^{n} \mathbb{I}_{\{\rho(X_i,X) < r_{k/n}(X)\}} Y_i - \sum_{i=1}^{n} \mathbb{I}_{\{X_i \in N_k(X)\}}Y_i \bigg| \nonumber \\
&= \frac{1}{k}\bigg|\sum_{i=1}^{n} \mathbb{I}_{\{X_i\in\Delta(X)\}} Y_i \bigg| \nonumber \\
&\leq \frac{1}{k}\bigg|\sum_{i=1}^{n} \mathbb{I}_{\{X_i\in\Delta(X)\}}\bigg| \nonumber \\
  & = \frac{1}{k} \bigg| \sum_{i=1}^{n} \mathbb{I}_{\{\rho(X_i,X) < r_{k/n}(X)\}}  -   \sum_{i=1}^{n} \mathbb{I}_{\{X_i \in N_k(X)\}} \bigg| \nonumber \ \\
 & = \bigg|\frac{1}{k} \sum_{i=1}^{n}  \mathbb{I}_{\{\rho(X_i,X) < r_{k/n}(X)\}}  -   1 \bigg|, \label{eqn:eta_bound}
 \end{align}
because $N_k(X)$ contains exactly $k$ points. 

Next we show that the latter term converges to zero.
Let $\hat{\eta}_{n}(X)$ be equal to $\frac{1}{k} \sum_{i=1}^{n}  \mathbb{I}_{\{\rho(X_i,X) < r_{k/n}(X)\}}$ and let $\hat{\eta}(X)$ be identically equal to $1$. 
Conditionally on $X=x$, the expected value of the random variable under the absolute sign is zero (LLN), which allows to pass to variance.
Using the Cauchy-Schwarz inequality, 
 \begin{align*}
 \mathbb{E}_{\tilde\mu^n}\{\mathbb{E}_{\mu}\{|{\eta}_n^*(X) - \eta_n(X)|\} \} & \leq \mathbb{E}_{\tilde\mu^n}\{\mathbb{E}_{\mu}\{|\hat{\eta}_n(X) - \hat{\eta}(X)|\} \} \  \\
 & \leq \mathbb{E}_{\mu}\bigg\{ \sqrt{\mathbb{E}_{\tilde\mu^n}\{|\hat{\eta}_n(X) - \hat{\eta}(X)|^2\}} \bigg\} \  \\
 & \leq \mathbb{E}_{\mu}\bigg\{ \sqrt{ \frac{n}{k^2}Var\{\mathbb{I}_{\{\rho(X_i,X) < r_{k/n}(X)\}} \} } \bigg\} \  \\
 & \leq \mathbb{E}_{\mu}\bigg\{ \sqrt{ \frac{n}{k^2}\mu(B(X,r_{k/n}(X))) }  \bigg\} \  \\
 & = \mathbb{E}_{\mu}\bigg\{ \sqrt{ \frac{n}{k^2}\frac kn } \bigg\} \\
&= \frac{1}{\sqrt{k}},
 \end{align*}
which term goes to zero as $k \rightarrow \infty$.

For the first term on the right hand side of Eq. (\ref{eq:triangle}),
\begin{align*}
 & \mathbb{E}_{\tilde\mu^n}\{\mathbb{E}_{\mu}\{|\eta(X) - {\eta}_n^*(X)|\} \} \ \\ & \leq  \mathbb{E}_{\tilde\mu^n}\{\mathbb{E}_{\mu}\{|\eta(X) - {\eta}_n(X)|\} \} + \mathbb{E}_{\tilde\mu^n}\{\mathbb{E}_{\mu}\{|\eta_n(X) - {\eta}_n^*(X)|\} \} \ \\
 & \leq  \mathbb{E}_{\mu}\{\mathbb{E}_{\tilde\mu^n}\{|\eta(X) - {\eta}_n(X)|\} \} + \mathbb{E}_{\tilde\mu^n}\{\mathbb{E}_{\mu}\{|\eta_n(X) - {\eta}_n^*(X)|\} \} \ \\
 & \rightarrow 0  \text{ as } n,k \rightarrow \infty, k/n \rightarrow 0,
 \end{align*}
where we used the fact that $\mathbb{E}_{\mu^n}\{|\eta(X) - \eta_n(X)| \} \rightarrow 0 $ because the $k$-NN rule in our setting is weakly consistent due to the results of Preiss and C\'erou--Guyader (Corollary \ref{c:preiss-cerou}).

The random variables $\abs{\eta(X) - {\eta}_n^*(X)}$ and $\abs{\hat{\eta}_n(X) - \hat{\eta}(X)}$ admit realisations as Borel measurable functions on $\Omega^{\infty}\times\{0,1\}^{\infty}\times\Omega$ taking values in $[0,1]$. Thus, the convergence in expectation implies convergence in measure, and consequently their restrictions to $\Omega^{\infty}\times\{0,1\}^{\infty}\times Q$, where by our assumption $Q\subseteq\Omega$ has a strictly positive measure, converge to zero as well, in measure and in expectation. So, for a given $\varepsilon > 0$ we can choose $n,k$ so large that
 \begin{align}
 \mathbb{E}_{\tilde\mu^n}\{\mathbb{E}_{\mu}\{|\eta(X) - {\eta}_n^*(X)|\mid X\in Q\} \} + \mathbb{E}_{\tilde\mu^n}\{\mathbb{E}_{\mu}\{|\hat{\eta}_n(X) - \hat{\eta}(X)|\mid X\in Q\} \} & < \frac{\varepsilon}{6}. \label{eqn:exp_1}
 \end{align}

Suppose we have random variables $X,Y$, such that $\E X+\E Y<\e_2$.
Let $\e_1>\e_2>0$. The event $X+Y>\e_1$ implies that either $\abs{X-\E X}$ or $\abs{Y-\E Y}$ strictly exceeds $(\e_1-\e_2)/2$. Indeed, assuming otherwise,
\[X+Y\leq \abs{X-\E X}+ \E X +\abs{Y-\E Y} +\E Y < \e_2 + (\e_1-\e_2) = \e_1.
\]

Writing
\begin{align*}
\mathbb{E}_{\mu}\left\{\left\vert\eta(X) - \eta_n(X)\right\vert \mid X\in Q \right\} &\leq 
\mathbb{E}_{\mu}\left\{\left\vert \eta(X) - \eta^\ast_n(X)\right\vert \mid X\in Q \right\} + 
\mathbb{E}_{\mu}\left\{\left\vert \eta^\ast_n(X) - \eta_n(X)\right\vert\mid X\in Q \right\}  \\
&\leq \mathbb{E}_{\mu}\left\{\left\vert \eta(X) - \eta^\ast_n(X)\right\vert\mid X\in Q \right\}  +
\mathbb{E}_{\mu}\left\{\left\vert \hat\eta_n(X) - \hat\eta(X)\right\vert \mid X\in Q \right\}
\end{align*}
(by (\ref{eqn:eta_bound})) and applying the above observaton with $\e_1=\varepsilon/2$ and $\e_2=\varepsilon/6$, we have
 \begin{align}
 &\mathbb{P}\bigg( \mathbb{E}_{\mu}\{|\eta(X) - \eta_n(X)|\mid X\in Q\} > \frac{\varepsilon}{2} \bigg) \nonumber \\ 
 & \leq \mathbb{P}\bigg( \mathbb{E}_{\mu}\{|\eta(X) - {\eta}_n^*(X)|\mid X\in Q\} - \mathbb{E}_{\tilde\mu^n}\{\mathbb{E}_{\mu}\{|\eta(X) - {\eta}_n^*(X)|\mid X\in Q\} \} > \frac{\varepsilon}{6} \bigg) +  \nonumber \\ & \ \ \ \  \mathbb{P}\bigg(\mathbb{E}_{\mu}\{|\hat{\eta}_n(X) - \hat{\eta}(X)| \mid X\in Q \}   - \mathbb{E}_{\tilde\mu^n}\{\mathbb{E}_{\mu}\{|\hat{\eta}_n(X) - \hat{\eta}(X)| \mid X\in Q\} \} > \frac{\varepsilon}{6} \bigg),  \label{eqn:bound_suc}
 \end{align}
where we used the inequality \eqref{eqn:exp_1}. Now we will separately estimate the probability of deviations in the two last terms.
 
For the first term let $\theta$ be a function defined on labeled samples, $\theta: (\Omega \times \{0,1\})^n \rightarrow [0,\infty)$ as
 \begin{align*}
 \theta(\sigma_n) = \mathbb{E}_{\mu}\{|{\eta}(X) - \eta_n^*(X)|\mid X\in Q \}.
 \end{align*}
 Let a new sample $\sigma_n^{'}$ be formed by replacing $(x_i,y_i)$ with $(\hat{x}_i,\hat{y}_i)$. 
The difference of values of ${\eta}_{ni}^*$ computed at the original sample and the altered one is at most $1/k$. For elements of $Q$, the value can only change at the points of the set $D(x_i,k/n)\cap Q$.
According to Lemma \ref{lem:stone_lemma_strong}, the $\mu$-measure of the latter set is bounded by $(\beta+1)k/n$ whenever $r_{k/n}$ is sufficiently small (smaller than the scale $s$, in fact). Therefore, the normalized (conditional) measure of this set in $Q$ is bounded by $(\beta+1)k/\mu(Q) n$, and 
 \begin{align}
 |\theta(\sigma_n) - \theta(\sigma_n^{'} ) |& \leq
\frac 1 k \cdot (\beta+1)\frac k{\mu(Q) n} \nonumber \\
&= \frac{\beta+1}{\mu(Q) n}. \label{eq:lipschitzkn}
 \end{align}

Let us remind a classical concentration inequality.

\begin{thrm}[Azuma, McDiarmid]
Let $X_1,X_2,\ldots,X_n$ be i.i.d. random variables taking values in a space $\Omega$, and let a function $f\colon \Omega^n\to \R$ satisfy the following Lipschitz condition with regard to the Hamming distance: whenever just the $i$-th coordinate in the argument $(x_1,x_2,\ldots,x_n)$ is changed, the value of the function changes by at most $c_i>0$. Then the probability of the deviation of the random variable $f(X_1,X_2,\ldots,X_n)$ from the expected value by at least $t>0$ is bounded by
\[2\exp\left(-\frac{2t^2}{\sum_{i=1}^nc_i^2} \right).\]
\label{th:azuma}
\end{thrm}

We conclude that 
\[\mathbb{P}\bigg( \mathbb{E}_{\mu}\{|\eta(X) - {\eta}_n^*(X)|\mid X\in Q\} - \mathbb{E}_{\tilde\mu^n}\{\mathbb{E}_{\mu}\{|\eta(X) - {\eta}_n^*(X)|\mid X\in Q\} \} > \frac{\varepsilon}{6} \bigg)\leq 2\exp\left(-\frac{\varepsilon^2\mu(Q)^2 n}{18(\beta+1)^2} \right).\]

An identical argument applied to $\hat{\eta}_n$ results in a similar concentration estimate for the second term in Eq. (\ref{eqn:bound_suc}), and we are done.

\section{\label{zero}Strong consistency in the non-archimedean case}

Here we show that the randomized tie-breaking approach to the $k$-NN classifier in the presence of distance ties adopted by Devroye, Gy\"{o}rfi, Krzy\.{z}ak, and Lugosi \cite{DGKL} (see our Subsection \ref{strongconsistency}) and used by them to prove the strong universal consistency of the $k$-NN classifier in the Euclidean setting works also in the case of metric spaces of non-Archimedian metric spaces: those whose metric satisfies the strong triangle inequality:
\[d(x,y)\leq \max\{d(x,z),d(z,y)\}.\]

However, the proof becomes somewhat trickier, revealing some interesting geometric features of non-Archimedian spaces with measure.

\begin{thrm}
The $k$-NN classifier is strongly universally consistent in every complete separable non-Archimedian metric space, under the tie-breaking strategy of Devroye, Gy\"{o}rfi, Krzy\.{z}ak, and Lugosi.
\label{th:strongnonarch}
\end{thrm}

\begin{rmrk}
A slightly more general class of metric spaces is formed by those of Nagata dimension zero: a metric space is non-Archimedian if and only if it has Nagata dimension zero on the scale $s=+\infty$, see \cite{CKP}, Example 5.3.
Our result above requires a minimal amount of adjustments to be extended to the complete separable metric spaces of Nagata dimension zero on some scale $s>0$. We decided to avoid technicalities in order to make the argument in the proof of Lemma \ref{l:4alpha} below a little clearer.
\end{rmrk}

\begin{figure}[hbt!] 
\begin{center}
  \scalebox{0.31}{\includegraphics{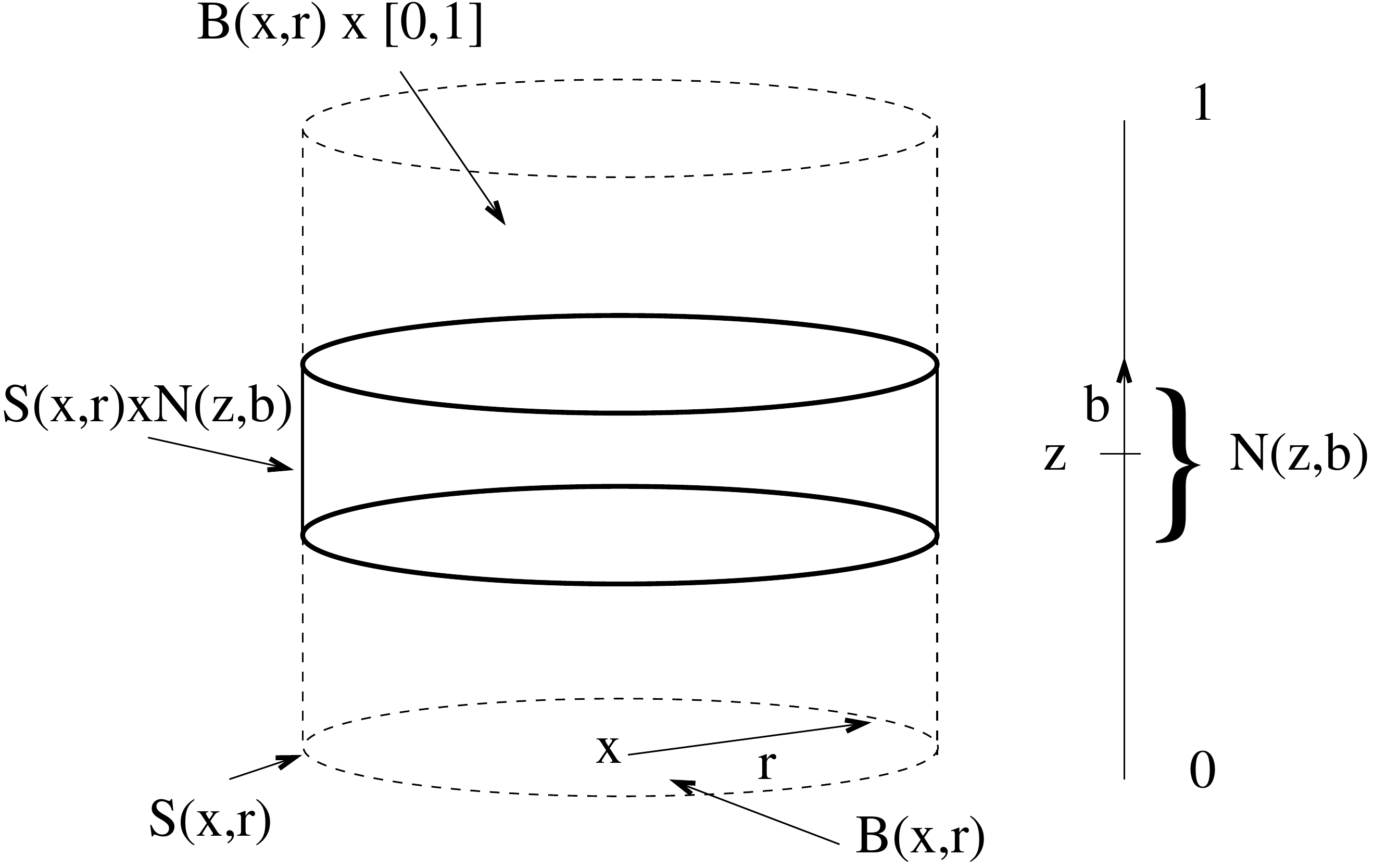}} 
  \caption{The set $B(x,z,r,b)=B(x,r)\times \I \cup S(x,r)\times N(z,b)$.}
  \label{fig:cylinder}
\end{center}
\end{figure}

We begin with combinatorial preparations. 
For $z\in\I$ and $b\geq 0$, denote
\[N(z,b)=\{x\in\I\colon \abs{z-x}\leq b\}.\]
Let $\Omega$ be a metric space.
Given $x\in\Omega$, $z\in\I$, $r,b>0$, define, just like in \cite{DGKL}, the set
\begin{equation}
B(x,z,r,b) = B(x,r)\times \I \cup S(x,r)\times N(z,b)\subseteq \Omega\times [0,1].
\label{eq:bxzrb}
\end{equation}
(See Fig. \ref{fig:cylinder}.)

Now let $\alpha>0$. Given $(x,z)\in\Omega\times\I$, denote $r_{\alpha}(x)$ as before (Eq. (\ref{eqn:r_alpha})), being the infimum of all the radii $r>0$ such that the open ball of radius $r$ around $x$ has measure $\geq\alpha$. It the presence of atoms, it may happen that $r=0$; we adopt the convention that $B(x,0)$ is the empty set, and $\bar B(x,0)=S(x,0)=\{x\}$.

\begin{lmm}
$\mu(B(x,r_{\alpha}(x))\leq\alpha$.
\end{lmm}

\begin{proof}
The statement is trivially true if $r_{\alpha}(x)=0$. Otherwise, approximate $r_{\alpha}(x)>0$ with a strictly increasing sequence  of radii $r_n\uparrow r_{\alpha}(x)$, and use sigma-additivity.
\end{proof}

Now define 
\begin{equation}
b_{\alpha}(x,z)=\inf\{b>0\colon \mu\otimes \lambda (B(x,z,r_{\alpha}(x),b))\geq \alpha\}.
\label{eq:balpha}
\end{equation}

\begin{lmm}
The function $b_{\alpha}\colon \Omega\times \I \to \R$ is Borel measurable.
\label{l:bborel}
\end{lmm}

\begin{proof}
One has
\[b_{\alpha}(x,z)= \begin{cases} b_{\alpha}(x,1/2),&\mbox{ if } b_{\alpha}(x,1/2)\leq z \leq 1-b_{\alpha}(x,1/2),\\
2b_{\alpha}(x,1/2)-\min\{z,1-z\} &\mbox{ otherwise.}
\end{cases}\]
Thus, it suffices to prove that $b_{\alpha}(x,1/2)$ is measurable as a function of $x\in\Omega$. This can be written as
\[b_{\alpha}(x,1/2) = \begin{cases} 
0,&\mbox{ if }\mu(S_{r_{\alpha}(x)}(x))=0, \\
\frac 12 (\alpha - \mu(B_{r_{\alpha}(x)}(x)) \mu(S_{r_{\alpha}(x)}(x))^{-1}&\mbox{ if }\mu(\bar B_{r_{\alpha}(x)}(x))>\alpha, \\
\frac 12,&\mbox { if }\mu(\bar B_{r_{\alpha}(x)}(x))=\alpha>\mu(B_{r_{\alpha}(x)}(x)).
\end{cases}
\]
Everything now reduces to proving the measurability of the maps $x\mapsto \mu\left(B_{r_{\alpha}(x)}(x)\right)$ and $x\mapsto \mu\left(\bar B_{r_{\alpha}(x)}(x)\right)$.

As the function $x\mapsto r_{\alpha}(x)$ is continuous (in fact, $1$-Lipschitz, see Lemma \ref{lem:r_alpha}), when $x$ is fixed and $x_n\to x$, we have $r_{\alpha}(x_n)\to r_{\alpha}(x)$. For every $\e>0$, from the triangle inequality, when $n$ is large enough, the closed ball $\bar B_{r_{\alpha}(x_n)}(x_n)$ is contained in the $\e$-neighbourhood of the ball $\bar B_{r_{\alpha}(x)}(x)$. When $\e\downarrow 0$, the measure of this $\e$-neighbourhood converges to the measure of $\bar B_{r_{\alpha}(x)}(x)$ by sigma-additivity of $\mu$, so we conclude
\[\limsup_{n\to\infty}\mu\left(\bar B_{r_{\alpha}(x_n)}(x_n)\right) \leq \mu\left(\bar B_{r_{\alpha}(x)}(x)\right).\]
Thus, the function $x\mapsto \mu\left(\bar B_{r_{\alpha}(x)}(x)\right)$ is upper semi-continuous, hence Borel measurable. An identical argument works for the sphere in place of the closed ball, and this suffices.
\end{proof}

For $r=r_{\alpha}(x)$ we have
\[\mu\otimes\lambda (B(x,r)\times \I )=\mu (B(x,r)) \leq \alpha \leq \mu(\bar B(x,r))= \mu\otimes\lambda (\bar B(x,r)\times \I ).\]
In case where the two values are different, the function
\[[0,1]\ni b \mapsto \mu (S(x,r))\times \lambda (N(z,b))\in [0,\mu(S(x,r))]\]
is continuous and surjective, so the value $\alpha$ is achieved. We have:

\begin{lmm}
For every $\alpha$, $0<\alpha<1$,
\[\mu\otimes\lambda (B((x,z,r_{\alpha}(x),b_{\alpha}(x,z)))=\alpha.\]
\label{l:remindthat}
\end{lmm}

Now, given $\alpha\in (0,1]$ and $(x,z)\in\Omega\times\I$, define
\begin{align*}
D(x,z,\alpha)&=\{(y,w)\in\Omega\times\I\colon (x,z)\in B(y,w,r_{\alpha}(y),b_{\alpha}(y,w)))\}\\
&=\{(y,w)\in\Omega\times\I\colon \mbox{ either }d(x,y)< r_{\alpha}(y)\mbox{ or }d(x,y)=r_{\alpha}(y)\mbox{ and }\abs{z-w}\leq b_{\alpha}(y,w)\}.
\end{align*}
The functions $d(x,y)$ and $\abs{z-w}$ with $x,z$ fixed are Lipschitz continuous with constant $1$, as is $r_{\alpha}(y)$ by Lemma \ref{lem:r_alpha}. The function $b_{\alpha}(y,w)$ is Borel measurable by Lemma \ref{l:bborel}. It follows that the set $D(x,z,\alpha)$, defined by inequalities involving those four functions, is Borel measurable.

The argument by Devroye et al. \cite{DGKL} was based on the following technical result (Lemma 3, {\em loco citato}): in the finite-dimensional Euclidean domain $\Omega=\R^d$, for every $x\in\Omega$, $z\in\I$ and $\alpha>0$, 
\begin{equation}
(\mu\otimes\lambda)(D(x,z,\alpha))\leq C\alpha,
\label{eq:Calpha}
\end{equation}
where $C=C(d)$ is a constant depending on the dimension of the space.

However, this kind of bound does not hold in more general finite-dimensional spaces in the sense of Nagata. In fact, it already fails in the non-archimedean metric spaces. Here is a counter-example.

\begin{xmpl}
\label{ex:strangebehaviour}
Let $\Omega$ be any infinite complete non-archimedean metric space having at least one non-isolated point. For example, one can take any of the classical examples such as the space of $p$-adic numbers $\Q_p$, or the Cantor space $\{0,1\}^\omega$ with the metric $d(x,y)=2^{-\min\{i\colon x_i\neq y_i\}}$. 

Fix a non-isolated point $x\in\Omega$ and a sequence $x_n$ converging to $x$, such that $r_n=d(x,x_n)$ is strictly decreasing. Denote $S_n=S_n(x_n,r_n)$ the spheres, $B_n= B(x_n,r_n)$ the open balls, and $\bar B_n=\bar B(x_n,r_n)$ the closed balls. 

For every $n$ we have the following, somewhat counter-intuitive, property, due to the strong triangle inequality:
\begin{equation}
\bar B_{n+1}\subseteq S_n.
\label{eq:ballsphere}
\end{equation}
Indeed, let $y\in \bar B_{n+1}$, that is, $d(x_{n+1},y)\leq r_{n+1}$. Then $d(x,y)\leq \max\{d(x,x_{n+1}),d(x_{n+1},y)\}= r_{n+1}$, because $x\in S_{n+1}$.
We have:
\[d(x_n,y)\leq\max\{d(x_n,x),d(x,y)\}=r_n,\]
and at the same time
\[r_n=d(x_n,x)\leq \max\{d(x_n,y),d(y,x)\}.\]
So we must have $d(x_n,y)=r_n$, proving Eq. (\ref{eq:ballsphere}).

In particular, the open balls $B_n$ are all two-by-two disjoint: if $n<m$, then $B_n\cap S_n=\emptyset$, while $B_m\subseteq S_n$. Also, the spheres $S_n$ form a nested sequence: $S_1\supseteq S_2\supseteq \ldots$.

Choose a probability measure $\mu$ on $\Omega$ so that for $n=1,2,\ldots$
\[\mu(B_n)=\frac 1n - \frac 1{n+1}=\frac{1}{n(n+1)}.\]
Then $\cup B_n$ has a full measure. Denote
\[s_n=\mu(S_{n})=\mu\left(\bigcup_{i= n+1}^{\infty}B_i\right) = \frac 1{n+1}.\]

Fix a sufficietly small $\alpha>0$, in the sense to be defined shortly. Denote $\zeta_n=\alpha (n+1)$. When $n\leq \alpha^{-1}-1$, we have $\zeta_n\in [0,1]$. If $y\in B_n$, then $d(x,y)=\max\{d(x,x_n),d(x_n,y)\}=r_n$. We have $B(y,r_n)=B(x_n,r_n)=B_n$. Therefore, for every $y\in B_n$ and $w\in [0,1]$,
\begin{align*}
(\mu\otimes\lambda)\left(B(y,w, r_n,\zeta_n )\right)&
\leq \mu(B_n) + 2\zeta_n \mu(S_{n}) \\
& = \frac{1}{n(n+1)} + 2\alpha \\
& \leq 3\alpha,
\end{align*}
whenever $n\geq \alpha^{-1/2}$.

Now let $x$ be the same non-isolated point as above, and set $z=0$. By the above reasoning, the set $D(x,z,3\alpha)$ contains every pair $(y,w)$ with $y\in B_n$ and $w\leq \zeta_n$, provided that
\[\alpha^{-1/2} \leq n \leq \alpha^{-1}-1.\]
When $\alpha>0$ is sufficiently small, we have 
\[\lceil \alpha^{-1/2}\rceil <\alpha^{-2/3}<\alpha^{-4/5}< \lfloor\alpha^{-1}\rfloor-1.\]

Since the balls $B_n$ are pairwise disjoint, we have
\begin{align*}
(\mu\otimes\lambda)\left(D(x,z,3\alpha)\right) &\geq \sum_{n=\lceil \alpha^{-1/2}\rceil}^{\lfloor \alpha^{-1}\rfloor -1} \zeta_n \mu(B_n) \\
&= \sum_{n=\lceil \alpha^{-1/2}\rceil}^{\lceil \alpha^{-1}\rceil -1}\alpha (n+1)\frac{1}{n(n+1)} \\
&=\alpha  \sum_{n=\lceil \alpha^{-1/2}\rceil}^{\lceil \alpha^{-1}\rceil-1}\frac 1{n} \\
&\geq \alpha \int_{\alpha^{-2/3}}^{\alpha^{-4/5}}\frac{dx}{x} \\
& = \alpha \left[-\frac 45\ln \alpha + \frac 23\ln \alpha \right]\\
&= -\frac{2\alpha\ln\alpha}{15},
\end{align*}
which expression is $\omega(\alpha)$ as $\alpha\to 0$. Thus, unlike in the Euclidean case, there is no upper bound on the size of the set $D(x,z,\alpha)$ that is linear in $\alpha$.
\end{xmpl}

However, this example does not contradict the strong consistency of the $k$-NN rule. Indeed, the proof of \cite{DGKL} proceeds as follows. The inequality (\ref{eq:Calpha}), taken with $\alpha=k/n$, implies, like in our earlier argument (Eq. (\ref{eq:lipschitzkn})), that the misclassification error of an auxiliary rule is a Lipschitz function on the cube $\Omega^n\times \{0,1\}^n$, equipped with the normalized Hamming distance, with $C$ being the Lipschitz constant. The Azuma inequality bounds the probability that the error deviates by more than $\varepsilon>0$ from the expectation by an expression of the form $2\exp(-\e^2 n /C^2)$, and the sequence of such upper bounds is summable, allowing one to use the Borel--Cantelli lemma. If we now assume that the upper bound on the size of $D(x,z,\alpha)$ is of the form $-\alpha\ln\alpha$, and substitute $\alpha=k/n$,
the Azuma inequality bounds the probability of a large deviation by something like $2\exp(-\e^2 n /(\ln n)^2)$. This sequence is still summable over $n$, so the Borel--Cantelli argument applies. 

It turns out that in fact the upper bound in the above example is (up to a constant) exact. 

\begin{lmm}
Let $\Omega$ be a non-Archimedian metric space equipped with a Borel probability measure $\mu$, let $x\in\Omega$, $z\in [0,1]$, and $\alpha>0$.
Then
\[(\mu\otimes\lambda)(D(x,z,\alpha))\leq 4\alpha(-\ln\alpha+1).\]
\label{l:4alpha}
\end{lmm}

\begin{proof}
Fix $\alpha>0$.
We will estimate the measure of the set $D(x,z,\alpha)\setminus (\{x\}\times \I)$. If $x$ is not an atom, this makes no difference. If $\mu\{x\}>0$, then for any pair of the form $(x,w)\in D(x,z,\alpha)$ the product measure of the set $\{x\}\times N(z,\abs{w-z})$ does not exceed $\alpha$. This means $\abs{z-w}\leq \alpha \mu\{x\}^{-1}$. Consequently, the measure of the set $D(x,z,\alpha)\cap (\{x\}\times \I)$ is bounded by $2\alpha$, and we will just add this value to our estimate at the end.

It simplifies things to estimate separately the measure of the intersection of the above set, $D(x,z,\alpha)\setminus (\{x\}\times \I)$, with $\supp\mu\times (0,z)$ and with $\supp\mu\times (z,1)$; as the arguments are identical, we will only do the former.

By approximating the measurable set $(D(x,z,\alpha)\setminus \{x\}\times \I)\cap \supp\mu\times (0,z)$ with a compact subset $K$ from inside to any given accuracy (Luzin's theorem), we can concentrate on bounding the measure of $K$. 

Denote $\mathcal V$ the family of all open subsets of $\Omega\times \I$ of the form $B(y,r)\times (w,z)$, where $(y,w)\in K$, $w<z$, and $r=d(y,x)$. Notice that they cover $K$. Choose a finite subcover of $K$ with sets of this form, say $B(y_i,r_i)\times (w_i,z)$, $i=1,2,\ldots,N$. Because the metric is non-Archimedian, we can assume these open balls, $B(y_i,r_i)$, to be disjoint from each other. (Indeed, assume $B(y_i,r_i)$ and $B(y_j,r_j)$ intersect, $i\neq j$. Then one of them is entirely contained in the other, say $B(y_i,r_i)\subseteq B(y_j,r_j)$, and as $r_i=d(y_i,x)=d(y_j,x)=r_j$ by the strong triangle inequality, we have $B(y_i,r_i)=B(y_j,r_j)$. Now out of the two sets $B(y_i,r_i)\times (w_i,z)$ and $B(y_j,r_j)\times (w_j,z)$, one contains the other, depending on whether $w_i$ or $w_j$ is smaller, so one can be discarded.)
Also, we can discard all balls of zero $\mu$-measure.
Order them in such a way that the radii $r_i$ decrease, and whenever $r_i=r_{i+1}$, we have $w_i\leq w_{i+1}$. 

Now, some more non-Archimedian geometry. For every $i$, $d(y_i,y_{i+1})=r_i$: it cannot be strictly smaller because the open balls $B(y_{i},r_{i})$ and $B(y_{i+1},r_{i+1})$ are disjoint, and cannot be strictly larger because both points are at a distance $\leq r_i$ from $x$. As a consequence, $B(y_{i+1},r_{i+1})\subseteq S(y_i,r_i)$. Indeed, if $y\in B(y_{i+1},r_{i+1})$, then $d(y_i,y)\leq\max\{d(y_i,y_{i+1}),d(y_{i+1},y)=r_i$, and the strict inequality is again impossible because the open balls do not meet. Notice that it is possible that $r_i=r_{i+1}$, in which case the closed ball $\bar B(y_{i+1},r_{i+1})$ will coincide with $\bar B(y_i,r_i)$. Write for short $B_i=B(y_i,r_i)$, $S_i=S(y_i,r_i)$. To sum up, the open balls $B_i$ are two-by-two disjoint, and if $i<j$, then $B_j\subseteq S_i$. Also, write $\xi_i=z-w_i$.

Denote $b_i=\mu(B_i)$, $i=1,2,\ldots$, and $s_i=\sum_{j=i+1}^N b_j$, $i=0,1,2,\ldots$. These $b_i$ and $s_i$ are all strictly positive by the choice of $K$.
Also, $s_i\leq\mu(S_i)$ for $i\geq 1$. For all $i=1,2,\ldots,N$, 
\[\mu (S_i)\times \xi_i \leq\alpha,\]
so in particular 
\[\xi_i\leq \frac{\alpha}{\mu(S_i)}\leq \frac{\alpha}{s_i}.\]
Thus,
\[b_i\xi_i=\mu(B_i)\xi_i\leq (s_{i-1}-s_i)\frac{\alpha}{s_i}=
\alpha\left(\frac{s_{i-1}}{s_i} -1 \right).\]
Denote 
\[\gamma_i=\frac{s_{i-1}}{s_i} -1=\frac{s_{i-1}-s_i}{s_i}.\]
We have $\gamma_i> 0$. Let $n\leq N$ be the largest integer satisfying $s_n\geq\alpha$. (If it does not exist, then $\mu(K)\leq\alpha$ and we are done.) Clearly,
\begin{align}
\mu\left(K\right) &\leq \sum_{i=1}^N \mu\otimes\lambda (B_i\times (w_i,z))
\nonumber\\
& \leq\sum_{i=1}^n b_i\xi_i + \mu\left(\cup_{i=n+1}^N B_i \right) 
\nonumber\\
&\leq\alpha \sum_{i=1}^n \gamma_i + \alpha.\label{eq:estt}
\end{align}
So it is enough to estimate $\sum_{i=1}^n\gamma_i$. With this purpose, write
\[\frac{s_i}{s_{i-1}}= 1 -\delta_i,\]
where $\delta_i> 0$. Thus,
\[\delta_i = 1- \frac{s_i}{s_{i-1}} = \frac{s_{i-1}-s_i}{s_{i-1}}
=\gamma_i\frac{s_i}{s_{i-1}},\]
and
\[\gamma_i = \delta_i\frac{s_{i-1}}{s_i}.\]
Also, $s_i=s_{i-1}(1-\delta_i)$, so
\[s_n=s_0\prod_{i=1}^n(1-\delta_i) \leq \prod_{i=1}^n(1-\delta_i).\]
Notice that if for some $i\leq N$ we have $s_{i-1}/s_i>2$, then
\[s_i<s_{i-1}-s_i=b_i=\mu(B_i)\leq\alpha,\]
meaning $i\geq n+1$. Therefore, $s_{i-1}/s_i\leq 2$ for all $i\leq n$, so
\[\gamma_i\leq 2\delta_i\]
for all $i=1,\ldots,n$. 

As for all $t\in [0,1)$, $\ln(1-t)\leq -t$, we get:
\begin{align*}
\ln\alpha&\leq\ln s_n\\
& \leq\ln\prod_{i=1}^n(1-\delta_i) \\
&= \sum_{i=1}^n\ln(1-\delta_i) \\
&\leq - \sum_{i=1}^n\delta_i,
\end{align*}
hence
\[\sum_{i=1}^n\delta_i\leq - \ln\alpha,\]
and 
\[\sum_{i=1}^n\gamma_i\leq -2\ln\alpha,\]
whence we get the estimate using (\ref{eq:estt}) and the remark at the start of the proof. 
\end{proof} 

\begin{rmrk}
The main result of this Section, Theorem \ref{th:strongnonarch}, would be established in the general case of a complete separable metric space sigma-finite dimensional in the sense of Nagata if we could verify the following.
Suppose a subspace $Q$ of a complete metric space $\Omega$ has Nagata dimension $\beta$ on a scale $s>0$ in $\Omega$. Is it true that, for some absolute constant $C>0$ and all sufficiently small $\alpha$,
\[\mu(D(x,z,\alpha)\cap Q)\leq -C(\beta+1)\alpha\ln\alpha?\]
Of course one could think of weaker estimates that will also suffice.
\label{r:general}
\end{rmrk}

We will model the proof of Theorem \ref{th:strongnonarch} on the proof of Theorem 1 in \cite{DGKL}. First, we remind that, by Lemma \ref{l:remindthat}, for every pair $(x,z)$ there is a unique pair $(r_{k/n}(x),b_{k/n}(x,z))$ defined as in Eqs. (\ref{eqn:r_alpha}) and (\ref{eq:balpha}) with $\alpha=k/n$. This leads us to define the regression function approximation 
\begin{align}
 \label{eqn:eta_star2}
 {\eta}_{n}^*(X,Z) = \frac{1}{k}\sum_{i=1}^{n} \mathbb{I}_{\{(X_i,Z_i)\in B(X,Z,r_{k/n}(X),b_{k/n}(X,Z))\}} Y_i.
 \end{align} 
We also have the regression function approximation
\[\eta_n(X,Z)=\frac{1}{k}\sum_{i=1}^{n} \mathbb{I}_{\{(X_i,Z_i) \in N_k(X,Z)\}}Y_i,\]
where the choice of the set $N_k(X,Z)$ of $k$ nearest neighbours of $X$ using the auxiliary random variable $Z$ is made using the same tie-breaking strategy as described at the beginning of the section.

We need to prove that, first, the difference $\eta(X)-\eta_n(X,Z)$ converges to zero in expectation (or in probability), which would mean the (weak) consistency of the algorithm, and second, for every $\e>0$ the probabilities of an $\e$-deviation of $\eta(X)-\eta_n(X,Z)$ from its expected value taken over all $n$-samples form a summable sequence. This will allow to apply the first Borel--Cantelli lemma and deduce the strong consistency.

We have, taking the expectation over random samples (that is, $\E$ stands for $\E_{\sigma\sim \mu^n\otimes\lambda^n}$),
\begin{align}
\vert\eta(X)-\eta_n(X,Z)\vert& \leq
\vert\eta(X)-\E{\eta}_{n}^*(X,Z)\vert \nonumber \\
& + \vert\E{\eta}_{n}^*(X,Z) - {\eta}_{n}^*(X,Z)\vert + 
\vert {\eta}_{n}^*(X,Z) - {\eta}_{n}(X,Z)\vert. \label{eq:3terms}
\end{align}

We will verify the convergence to zero in expectation for all three terms. As to the deviation bound, the first term does not depend on a random sample, so the $\e$-deviation is improbable. We will deduce a summable bound for the second term, while the conclusion for the third will come for free as a particular case.

Notice that whenever a metric space with measure satisfies the strong Lebesgue--Besicovitch property (Eq. \ref{eq:lbs}), that is, for a.e. $x$,
\begin{equation}
\frac{1}{\mu(B(x,r))}\int_{B(x,r)} f(x)\,d\mu(x) \to f(x),
\label{eq:lb2}
\end{equation}
one can use closed balls in place of open balls and the a.e. convergence will still take place. Indeed, as every closed ball is the intersection of a sequence of open balls of the same centre, we have, by sigma-additivity,
\[\frac{1}{\mu(\bar B(x,r))}\int_{\bar B(x,r)} f(x)\,d\mu(x)=
\lim_{\e\downarrow r}\frac{1}{\mu(B(x,\e))}\int_{B(x,\e)} f(x)\,d\mu(x),\]
from where the statement follows. 

Let now $f\colon\Omega\to\R$ be an $L^1(\mu)$-function. By the main theorem of Preiss from \cite{preiss83} (reproduced above as Theorem \ref{th:preiss}), combined with our  observation in the previous paragraph, almost every $x\in\Omega$ has the property: given $\e>0$, one can select $\rho>0$ so small that when $0<r<\rho$, then the average value of $f$ in the $r$-ball around $x$, either open or closed, is $\e$-close to $f(x)$. 
We can also see $f$ as a function on $\Omega\times\I$ which only depends on the first argument, $x\in\Omega$. Now let $x$, $\e>0$, and $r$ are as above, and $z,\xi\in [0,1]$. Denote provisionally 
\[\overset{t}{B}(x,r) = \{y\in\Omega\colon (y,t)\in B(x,z,r,\xi)\} =
\begin{cases} \bar B(x,r),&\mbox{ if }\abs{z-t}\leq\xi,\\
B(x,r),&\mbox{ otherwise}
\end{cases}
\]
the ``horizontal section'' of $B(x,z,r,\xi)$ at the height $t\in [0,1]$. Now
the Fubini theorem implies
\begin{align*}
\int_{B(x,z,r,\xi)}\left\vert f(y)-f(x)\right\vert\,d\mu(y)d\lambda(t) & =
\int_0^1 d\lambda(t)\int_{\overset{t}{B}(x,r)}\left\vert f(y)-f(x)\right\vert\,d\mu(y) \\
&< \int_0^1 \e \mu(\overset{t}{B}(x,r))\,d\lambda(t)\\
&= \e (\mu\otimes\lambda)B(x,z,r,\xi).
\end{align*}
Thus, for $\mu$-a.e. $x\in\Omega$, the average value of $f$ over $B(x,z,r,\xi)$ converges to $f(x)$ when $r\downarrow 0$. In particular, this conclusion applies to the regression function $\eta$ and its average value over $B(x,z,r_{k/n}(x),b_{k/n}(x,z))$ when $n,k\to\infty$ and $k/n\to 0$.

It follows that the expected value $\E_{\sigma\sim \mu^n\otimes\lambda^n}{\eta}_{n}^*(x,z)$ of the approximation $\eta_n^{\ast}(x,z)$ taken over all random labelled $n$-samples converges to $\eta(x)$ for a.e. $x,z$ as $n,k\to\infty$ and $k/n\to 0$:
\begin{align*}
\E_{\sigma\sim \mu^n\otimes\lambda^n}{\eta}_{n}^*(x,z)&= \frac{1}{\mu\otimes\lambda(B(x,z,r_{k/n}(x),b_{k/n}(x,z)))}
\int_{B(x,z,r_{k/n}(x),b_{k/n}(x,z))} \E(Y\mid X=x^\prime,Z=z^\prime)\,d\mu(x^\prime)d\lambda(z^\prime) \\
&\to \E(Y\mid X=x,Z=z) \\
&= \eta(x).
\end{align*}

By the dominated convergence theorem, the integral of the first term in Eq. (\ref{eq:3terms}) over our extended domain, $\Omega\times\I$, converges to zero:
\[\E_{\mu\otimes\lambda} \vert\eta(X)-\E_{\tilde\mu^n}{\eta}_{n}^*(X,Z)\vert \to 0.\]

For the second term in Eq. (\ref{eq:3terms}) we use the argument already seen in the proof of Lemma \ref{l:main}. If the labelled sample is changed in one labelled point, then the value of 
\[\vert\E_{\tilde\mu^n}{\eta}_{n}^*(x,z) - {\eta}_{n}^*(x,z)\vert\]
may change by at most $1/k$, and only on a set of points $(x,z)$ having measure at most $4(k/n)(\ln n-\ln k +1)$, thanks to Lemma \ref{l:4alpha}. Therefore, the integral
\[\int \vert\E_{\tilde\mu^n}{\eta}_{n}^*(x,z) - {\eta}_{n}^*(x,z)\vert\,d(x,z)\]
changes its value by at most $(4/n)(\ln n-\ln k +1)\leq 4\ln n/n$. The Azuma--McDiarmid inequality (Theorem \ref{th:azuma}) implies that
\begin{align*}
\left\vert \int \vert\E{\eta}_{n}^*(x,z) - {\eta}_{n}^*(x,z)\vert\,d(x,z) -
\E\int \vert\E{\eta}_{n}^*(x,z) - {\eta}_{n}^*(x,z)\vert\,d(x,z)\right\vert 
\leq 2\exp\left(-\frac{\e^2 n}{8(\ln n)^2} \right).
\end{align*}
This is a summable sequence.

For the second term in Eq. (\ref{eq:3terms}) it remains to show convergence to zero in expectation. We perform a familiar trick with the Cauchy--Schwarz inequality and the variance:
\begin{align*}
\E_{\tilde\mu^n}\int \vert\E_{\tilde\mu^n}{\eta}_{n}^*(x,z) - {\eta}_{n}^*(x,z)\vert\,d(x,z) &
\leq \int\sqrt{\E_{\tilde\mu^n}\,\vert \E_{\tilde\mu^n}{\eta}_{n}^*(x,z) - {\eta}_{n}^*(x,z)\vert^2}\,d(x,z)\\
&\leq \int \sqrt{
\frac{1}{k^2}n\mathrm{Var}\left(Y\mathbb{I}_{(X,Z)\in B(x,z,r_{k/n}(x),b_{k/n}(x,z))} \right)
}d(x,z) \\
&\leq \int \sqrt{\frac{1}{k^2}n\mu\otimes\lambda(B(x,z,r_{k/n}(x),b_{k/n}(x,z))
}d(x,z) \\
&\leq \int \sqrt{\frac{1}{k^2}n\frac k n }d(x,z) \\
&=\sqrt{\frac{1}{k}}\to 0.
\end{align*}

Now the third term in Eq. (\ref{eq:3terms}). Let $(X_{(k)},Z_{(k)})$ denote the $k$-th nearest neighbour of $X$ in the random sample (in the order defined by the adopted tie-breaking). Denote $R_n=d(X,X_{(k)})$ and $B_n=\abs{Z-Z_{(k)}}$. Then
\begin{align}
\vert {\eta}_{n}^*(X,Z) - {\eta}_{n}(X,Z)\vert &=
\frac{1}{k}\left\vert\sum_{i=1}^{n} \mathbb{I}_{\{(X_i,Z_i)\in B(X,Z,r_{k/n}(X),b_{k/n}(X,Z))\}} Y_i - 
\sum_{i=1}^{n} \mathbb{I}_{\{(X_i,Z_i)\in B(X,Z,R_n,B_n)\}}Y_i\right\vert\nonumber \\
&\leq \frac{1}{k}\sum_{i=1}^{n}
\left\vert \mathbb{I}_{\{(X_i,Z_i)\in B(X,Z,r_{k/n}(X),b_{k/n}(X,Z))\}} -\mathbb{I}_{\{(X_i,Z_i)\in B(X,Z,R_n,B_n)\}} \right\vert\nonumber \\
&= \frac{1}{k}\left\vert\sum_{i=1}^{n} \mathbb{I}_{\{(X_i,Z_i)\in B(X,Z,r_{k/n}(X),b_{k/n}(X,Z))\}} - 1\right\vert.
\label{eq:thelast}
\end{align}
We have used the following three observations: the empirical measure of the symmetric difference of the sets $B(X,Z,r_{k/n}(X),b_{k/n}(X,Z)$ and $B(X,Z,R_n,B_n)$ bounds the error, for the latter set this empirical measure is always one, and among the two intersections of the sample with these sets one always contains the other. Now introduce the regression function $\hat\eta\equiv 1$ and the corresponding approximation 
\[\hat\eta^{\ast}=\frac{1}{k}\sum_{i=1}^{n} \mathbb{I}_{\{(X_i,Z_i)\in B(X,Z,r_{k/n}(X),b_{k/n}(X,Z))\}}.\]
The last line of the equation (\ref{eq:thelast}) becomes $\left\vert \hat\eta^{\ast} - \E \hat\eta^{\ast}\right\vert$,
and is therefore just a special case of the second term correspoding to the constant regression function $\hat\eta\equiv 1$.

\section{\label{s:conjecture}The revised conjecture}

We propose the following conjecture (a revised version of the conjecture previously stated by us in \cite{CKP}).

\begin{cnjctr}
For a complete separable metric space $\Omega$, the following are equivalent.
\begin{enumerate}
\item The $k$-NN classifier is (weakly) universally consistent in $\Omega$.
\item For every sigma-finite locally finite Borel measure $\mu$ on $\Omega$, every $L^1(\mu)$-function $f\colon\Omega\to\R$ satisies the weak Lebesgue--Besicovitch differentiation property: 
\begin{equation}
\frac{1}{\mu(B(x,r))}\int_{B(x,r)} f(x)\,d\mu(x) \to f(x)
\end{equation}
in probability.
\item The space $\Omega$ is sigma-finite dimensional in the sense of de Groot, that is, one can represent $\Omega$ as a union of subspaces $W_n$ in such a way that for each $n$ and some $\delta_n\in\N$ and $s_n>0$, every finite family of closed balls with centres in $W_n$ having the same radii $<s_n$ admits a subfamily covering all the centres of the original balls and having multiplicity $\leq\delta_n+1$ in $\Omega$.
\end{enumerate}
\end{cnjctr}

The implication (3)$\Rightarrow$(2) follows from the results of \cite{AQ}, and the implication (2)$\Rightarrow$(1) was established in \cite{CG}. Thus, only (1)$\Rightarrow$(3) needs to be verified.

\section*{Acknowledgement}
We are most grateful to the anonymous ESAIM:PS referee who has read with utmost care both the original version of the article and the subsequent major revision of it, and whose reports permitted us to improve the paper very considerably, in particular to discover Example \ref{ex:strangebehaviour}, state the correct version of Lemma \ref{l:4alpha}, and clarify the proof of Proposition \ref{p:notstronglyconsistent}. Of course the remaining errors and obscure passages are all authors' own.

\end{document}